\documentclass[10pt]{article}
\usepackage[a4paper, total={6in, 8in}]{geometry}

\usepackage[utf8]{inputenc}
\usepackage{graphicx}
\usepackage{amsmath}
\usepackage{amssymb}
\usepackage{url}
\usepackage{caption}
\usepackage[sort, numbers]{natbib}
\usepackage[dvipsnames]{xcolor}
\usepackage[section]{placeins} 
\usepackage{multirow}
\usepackage[normalem]{ulem}
\usepackage{authblk}
\usepackage[labelformat=empty, position=top]{subcaption}
\usepackage[export]{adjustbox}
\usepackage{xr}
\usepackage{tikz}
\usepackage{etoolbox}

\usepackage[absolute]{textpos}

\makeatletter
\newcommand*{\addFileDependency}[1]{
  \typeout{(#1)}
  \@addtofilelist{#1}
  \IfFileExists{#1}{}{\typeout{No file #1.}}
}
\makeatother

\newcommand*{\myexternaldocument}[1]{%
    \externaldocument{#1}%
    \addFileDependency{#1.tex}%
    \addFileDependency{#1.aux}%
}

\newrobustcmd*{\uptriangle}[1]{\tikz{\filldraw[draw=#1,fill=#1] (0,0) -- (0.2cm,0) -- (0.1cm,0.2cm);}}
\newrobustcmd*{\downtriangle}[1]{\tikz{\filldraw[draw=#1,fill=#1] (0,0.2cm) -- (0.2cm,0.2cm) -- (0.1cm,0);}}

\myexternaldocument{supplementary}

\begin{document}

\begin{textblock}{10}(3,1)
\noindent\Large This work has been submitted to the IEEE for possible publication. Copyright may be transferred without notice, after which this version may no longer be accessible.
\end{textblock}

\title{Uncertainty-aware deep learning methods for robust diabetic retinopathy classification}
\author[1]{Joel~Jaskari}
\author[1]{Jaakko~Sahlsten}
\author[2,3]{Theodoros~Damoulas}
\author[4]{Jeremias~Knoblauch}
\author[5]{Simo~Särkkä}
\author[5]{Leo~Kärkkäinen}
\author[6]{Kustaa~Hietala}
\author[1,2,*]{Kimmo~Kaski}

\affil[1]{Department of Computer Science, Aalto University, Finland}
\affil[2]{The Alan Turing Institute, London, United Kingdom}
\affil[3]{University of Warwick, Coventry, United Kingdom}
\affil[4]{University College London, London, United Kingdom}
\affil[5]{Department of Electrical Engineering and Automation, Aalto University, Finland}
\affil[6]{Central Finland Health Care District, Finland}
\affil[*]{Corresponding author}

\maketitle

\begin{abstract}
Automatic classification of diabetic retinopathy from retinal images has been widely studied using deep neural networks with impressive results. However, there is a clinical need for estimation of the uncertainty in the classifications, a shortcoming of modern neural networks. Recently, approximate Bayesian deep learning methods have been proposed for the task but the studies have only considered the binary referable/non-referable diabetic retinopathy classification applied to benchmark datasets. We present novel results by systematically investigating a clinical dataset and a clinically relevant 5-class classification scheme, in addition to benchmark datasets and the binary classification scheme. Moreover, we derive a connection between uncertainty measures and classifier risk, from which we develop a new uncertainty measure. We observe that the previously proposed entropy-based uncertainty measure generalizes to the clinical dataset on the binary classification scheme but not on the 5-class scheme, whereas our new uncertainty measure generalizes to the latter case.

\end{abstract} 

\section{Main}

Deep neural networks have achieved impressive results in a wide variety of problems, ranging from large scale image classification \citep{efficient}, to natural language understanding \citep{gpt3}, and medical image segmentation \citep{ronneberger+15}. However, the standard methods have been found to produce over-confident predictions, meaning that they are poorly calibrated \citep{calibrationmodern}. In classification tasks, a poorly calibrated network can place a high probability on one of the classes, even when the predicted class is incorrect, whereas a well calibrated classifier would place less probability mass on uncertain classes. The issue of uncertainty estimation is especially important in the medical domain, in order to trust confident model predictions for screening automation and referring uncertain cases for manual intervention of a medical expert. In this work we refer to the classifiers that can indicate their uncertainty as robust classifiers.

Over the past few years, the automatic classification of diabetic retinopathy by using deep neural networks has been under growing interest \citep{Abramoff2016,gulshan,ting,sahlsten+19}. More recently, the focus of attention has turned on developing robust deep learning methods for the classification task, most commonly using the approximate Bayesian deep learning approach that approximates the Bayesian neural network (BNN) posterior distribution in a computationally scalable manner. The previous works have considered a variety of aspects from studying the benefits of uncertainty estimates \citep{nature_leveraging} to algorithmic development of robust methods \citep{radial,auguncertainty}. Although the variety of different studied algorithms has been
diverse \citep{nature_leveraging,systematic}, the used datasets have been benchmark datasets. This leaves open the question of whether the algorithms generalize to clinical data. In addition, these recent works have mainly focused on the classification of diabetic retinopathy using binary classification schemes, i.e. "referable vs. non-referable" (RDR) or "healthy vs. any" diabetic retinopathy. However, in clinically oriented approaches there has been a shift towards the 5-class proposed international diabetic retinopathy classification system (PIRC) \citep{Krause2018,Ruamviboonsuk2019,sahlsten+19}. In order for these approaches to have clinical use, it is of paramount importance that the algorithms generalize to clinical datasets and diabetic retinopathy grading systems, that remain yet unstudied. 

The common aspect among the works focusing on robust methods is the use of uncertainty information to simulate a referral process, introduced by \citet{nature_leveraging}. Each prediction is associated with an uncertainty estimate, and the least certain predictions are referred to experts, while the more certain predictions are used for evaluation. This process mimics a situation in which the automated system asks human intervention for uncertain cases, i.e. refers them to an expert. In practice, the holdout test set predictions are ordered according to their uncertainty and several referral levels are defined corresponding to a percentile of referred examples, i.e. 10\% referral level means that 10\% of the most uncertain examples are left out of the evaluation.

In \citet{nature_leveraging}, a Monte Carlo (MC) dropout neural network was used for two binary classification tasks: classification of any diabetic retinopathy and RDR. They used the EyePACS dataset \citep{eyepacs} for training and testing and evaluated the out-of-distribution performance with the Messidor dataset \citep{messidorcite1}. They observed improved robustness in comparison to a baseline standard neural network.
\citet{systematic} conducted a more methodologically extended study for classification of RDR. The EyePACS dataset was used for training and testing, and out-of-distribution performance was evaluated with the APTOS \citep{aptosdataset}. The work examined a number of approximate Bayesian deep learning methods, such as the MC dropout, Mean Field Variational Inference (MFVI), deep ensembles, and MC dropout ensemble. They observed that the approximate Bayesian methods outperformed the standard neural network in all the experiments where network uncertainty was utilized. 
\citet{radial} proposed the Radial BNN method. The model was benchmarked against MFVI, MC dropout, and deep ensembles using the EyePACS dataset, as both the training and test sets, for the RDR classification task. A single Radial BNN was found to outperform the MC dropout and MFVI, but not the deep ensemble. Overall, an ensemble of Radial BNN's turned out to outperform all other methods on all referral levels.
In \citet{galnew} the RDR task was examined using the EyePACS dataset for training and testing, and out-of-distribution performance was evaluated with the APTOS, similar to \citet{systematic}. They also studied how the robust deep learning methods generalize when no images of severe and proliferative diabetic retinopathy are used in the training. The considered approximate deep learning methods were MFVI, Radial BNN, Function-Space Variational Inference, MC dropout, and Rank-1 Parameterized BNN, and ensembles of them. Also deep ensembles were used. It was found that the MC dropout ensemble performed the best for the within-distribution and the MFVI ensemble for the out-of-distribution experiments.

In this study, our objective is to analyze robust neural networks, i.e. networks that are inherently well calibrated, for the task of diabetic retinopathy classification on both RDR and PIRC classifications. For this reason we leave out the analysis of \textit{post-hoc} methods, such as the test-time augmentation introduced in \citet{auguncertainty} and \citet{ayhan2020expert}, neural network softmax temperature scaling, and probability binning strategies \citep{calibrationmodern}.

Many measures of uncertainty have been used in the previous works. In \citet{nature_leveraging}, the standard deviation of the output of the BNN and the entropy of the posterior predictive distribution were considered as measures of uncertainty and found to perform similarly. In \citet{systematic}, \citet{ayhan2020expert}, and \citet{galnew} the entropy was also selected as the measure of uncertainty. On the other hand, the mutual information between the parameters of the model and the output was considered as the measure of uncertainty in \citet{radial}.

In the present work, we explore the benefits of uncertainty estimates for a clinical dataset of a Finnish hospital on both the binary RDR and the 5-class PIRC classification schemes. We investigate 9 different approximate Bayesian methods to extensively analyze recently proposed methods. In addition, we study the out-of-distribution performance using three benchmark datasets: the EyePACS \citep{eyepacs}, the Messidor-2 \citep{messidorcite1,messidorcite2}, and the APTOS \citep{aptosdataset}. We observe that the entropy uncertainty estimates improve AUC performance in the binary RDR system, but in the case of the 5-class PIRC system, the QWK only improves across the benchmark datasets. For the clinical dataset and PIRC system, we observe less robust classifier performance using entropy based uncertainty. To improve the quality of the uncertainty estimates, we additionally propose a novel classifier risk based uncertainty measure, which improves the within-distribution uncertainty performance on both the Finnish hospital dataset and the EyePACS dataset. As far as we know, clinical diabetic retinopathy severity schemes and clinical workflow datasets have not been studied before using robust neural networks.

\section{Results}
\subsection{Approximate Bayesian methods}
Here we consider 9 different BNN approaches. The methods approximate the true distribution of the neural network parameters with different mechanisms. These methods are deep ensembles \citep{ensemble}, Monte Carlo (MC) dropout \citep{mcdropout}, Mean Field Variational Inference (MFVI) \citep{graves}, Generalized Variational Inference (GVI) \citep{gvi}, and Radial BNN \citep{radial}. We use the ensembling approach also for MC dropout, MFVI, GVI, and Radial BNN. Our baseline is a standard neural network trained with dropout and L2-regularization, which is equivalent to the maximum a posteriori (MAP) estimate if a fully factorized normal prior, with the variance inversely proportional to the regularization weight, is used on the network parameters.

The BNNs produce an estimate for the posterior distribution, which can be used to take the uncertainty in the parameters into account when making a prediction \citep{galthesis}. To measure the uncertainty, we use the entropy of the posterior predictive distribution, which has also been used in previous works. We show that this measure optimizes the negative log-likelihood when used in the referral process. This is because the referral process is a type of reject option classification with the uncertainty used as a risk measure. We propose a novel class of uncertainty measures based on the risk interpretation, from which we derive the QWK-Risk measure used as another uncertainty measure for the 5-class PIRC system.

The utility of the uncertainty information is evaluated in a similar manner as in previous works. We compute the uncertainty as the entropy or as the QWK-Risk of the posterior predictive distribution and reject some proportion of the most uncertain cases, which is called the referral level. The performance is evaluated for the binary RDR task using the area under the receiver operating characteristic curve (AUC), and for the 5-class PIRC task we use the quadratic weighted Cohen's Kappa (QWK), similar to \citet{Krause2018} and \citet{Ruamviboonsuk2019}. Both the AUC and QWK are evaluated at 0\% (no referral), 30\%, and 50\% referral levels, and are presented in a scale with maximum of 100 to improve readability.

\subsection{Datasets}
We use the following four datasets for our experiments: the EyePACS \citep{eyepacs}, KSSHP\citep{retinaduodecim}, Messidor-2 \citep{messidorcite2}, and APTOS \citep{aptosdataset}. The EyePACS and APTOS datasets were introduced for two different Kaggle competitions of diabetic retinopathy detection. The Messidor-2 is a common benchmark dataset that was introduced for research in computer-assisted diagnosis of diabetic retinopathy. The EyePACS, APTOS, and Messidor-2 datasets have been widely used in literature for training and analyzing robust neural networks. In order to examine if the results generalize to clinical hospital datasets, we use the non-public KSSHP dataset. The KSSHP set was collected from clinical workflow data from the Central Finland Health Care District. All the used datasets originate from different countries: The EyePACS from USA, KSSHP from Finland, APTOS from India, and Messidor-2 from France, allowing for extensive analysis for the generalization of the models under distribution shift by country.

\begin{figure}[tb]
    \centering
 \begin{subfigure}[t]{0.03\textwidth}
    \textbf{A.}
  \end{subfigure}
  \begin{subfigure}[t]{0.29\textwidth}
    \includegraphics[width=\linewidth, valign=t]{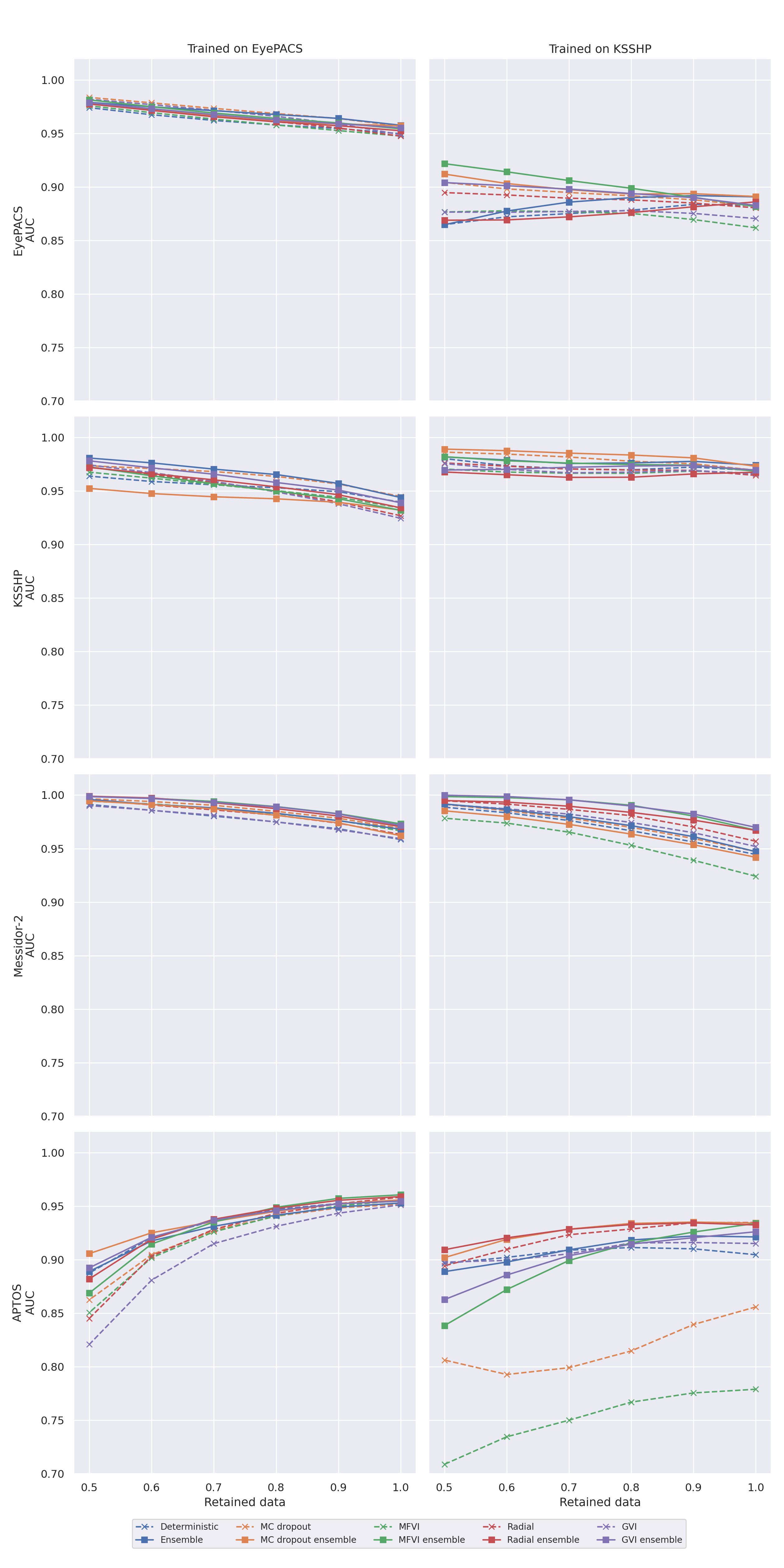}  
    \end{subfigure}\hfill
 \begin{subfigure}[t]{0.03\textwidth}
    \textbf{B.}
  \end{subfigure}
  \begin{subfigure}[t]{0.29\textwidth}
    \includegraphics[width=\linewidth, valign=t]{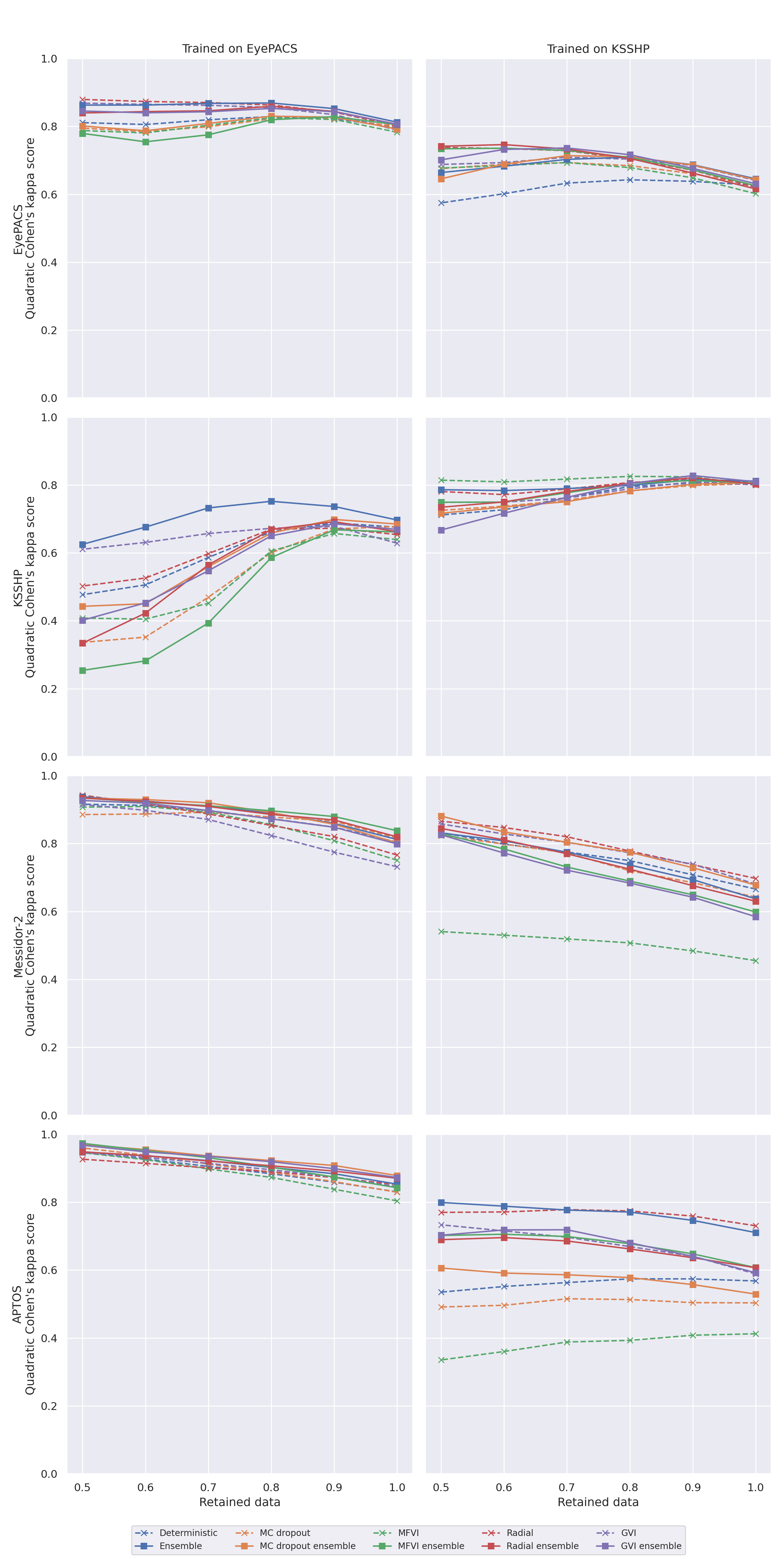}
    \end{subfigure}\hfill
 \begin{subfigure}[t]{0.03\textwidth}
    \textbf{C.}
  \end{subfigure}    
  \begin{subfigure}[t]{0.29\textwidth}
    \includegraphics[width=\linewidth, valign=t]{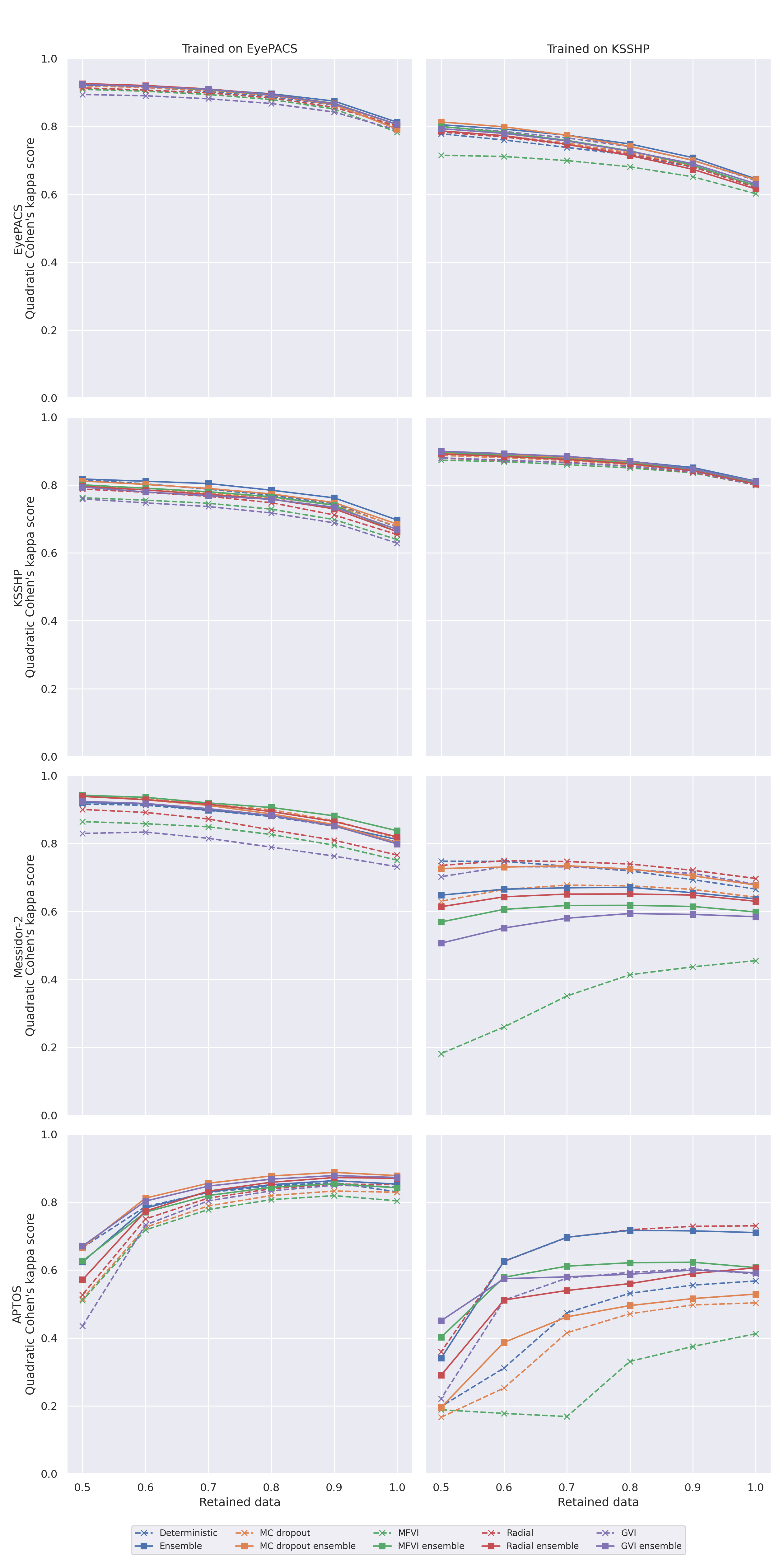}
\end{subfigure}\hfill
    \caption{Retained performance using (A) posterior predictive entropy to refer on the RDR task, (B) posterior predictive entropy to refer on the PIRC task, and (C) QWK-Risk to refer on the PIRC task. Each column shows which dataset was used for training and each row which test dataset was used for the results.}
    \label{fig:retained}
\end{figure}

\subsection{RDR classification}
Visual illustration of results for retained performance for the models trained on RDR classification is presented in Figure \ref{fig:retained}A. Detailed RDR results are presented in Table \ref{tab:rdr_entropy}. On the full test set, i.e. 0\% referral level, our EyePACS data trained models had better results on all of the benchmark sets in comparison to previous works, thus setting a higher standard as our baseline performance. Indeed, our worst models are better than the best models in previous works: on EyePACS \citet{nature_leveraging} had 92.7 AUC, \citet{systematic} 82.5 AUC, \citet{radial} 94.5 AUC, and \citet{galnew} 92.5 AUC, in comparison to our, MFVI and Radial, with 94.8 AUC. In addition, on APTOS dataset \citet{systematic} had 77.2 AUC and \citet{galnew} 94.6 AUC, whereas our worst models, MC dropout and GVI, had 95.2 AUC. The best Messidor result in \citet{nature_leveraging} was 95.5 AUC, whereas our worst model, deterministic MAP network, had 95.9 AUC, however, our results are not directly comparable since they use the Messidor set, which is a subset of the more recent Messidor-2 set.

Also, all of the KSSHP data trained models had a high AUC on 0\% referral level for the KSSHP test set. Indeed, every model outperforms the corresponding model trained and tested on the EyePACS dataset. However, the generalization of the KSSHP trained models to Messidor-2 and APTOS sets was worse in every comparison than the generalization of the EyePACS trained models.

The uncertainty enabled referral process results for the EyePACS data trained models are similar to \citet{nature_leveraging}, as we observe the referral process to systematically increase the AUC on the EyePACS test set and on the Messidor-2 set, and as a novel result, we observe that the referral process also increases the clinical KSSHP dataset AUC. We observe different behaviour on the APTOS dataset than in \citet{systematic} but similar to some models in \citet{galnew}, as the AUC of all models systematically decreases when utilizing the uncertainty. Surprisingly, we observe strong uncertainty estimates for the deterministic MAP network, unlike in \citet{nature_leveraging} and \citet{systematic}, that observed generally worse uncertainty estimates for the point estimate approach. This difference is likely due to the strong regularization we used in training our MAP network.

It turns out that the models trained with the KSSHP data had different utility from the uncertainty referral than those trained using the EyePACS set. Specifically, on the KSSHP test set, from the 0\% to 30\% referral level, the MFVI and Radial ensemble models exhibit slightly decreased performance. In contrast, the GVI model performance stayed constant across this range, but decreased from 30\% to 50\%. In addition, the uncertainty estimates do not generalize to the EyePACS test set to such an extent as they did using the EyePACS set trained models on the KSSHP set. We see that on the EyePACS set, the uncertainty estimates given by the MAP network, deep ensemble, and the Radial ensemble cannot be used to improve the performance by referral, and the MFVI and GVI models do not improve upon the 30\% referral rate. However, all the models perform well to the Messidor-2 dataset on all referral levels with the GVI ensemble even attaining 100.0 AUC for the 50\% referral level. In the case of APTOS dataset we observe similar performance as with the EyePACS trained model, as the performance decreases along the referral rate, with the exception of baseline network improving from 0\% to 30\%.

In terms of overall performance, we observe that best results for the EyePACS, KSSHP, and the Messidor-2 datasets were obtained by referring data. For EyePACS and KSSHP the best results were for within-distribution trained models. The best performance on the EyePACS was obtained with MC dropout model that reaches 98.4 AUC with 50\% referral level and on the KSSHP with MC dropout ensemble that reached 98.9 AUC with 50\% referral level. The best out-of-distribution performance on the Messidor-2 set was obtained with the KSSHP trained GVI ensemble with 50\% referral level that reached 100.0 AUC. For the APTOS set the referral did not improve performance, as the EyePACS trained MFVI ensemble with 0\% referral level reached the highest AUC of 96.1.

\begin{table}[h]
\centering
\makebox[\textwidth]{
\resizebox{\textwidth}{!}{%
\begin{tabular}{c|c|l|l|l||l|l|l}
 \multirow{2}{*}{Test Dataset} & \multirow{2}{*}{Method} & \multicolumn{3}{c||}{EyePACS trained} & \multicolumn{3}{c}{KSSHP trained} \\
 & & AUC Ref. 50\% & AUC Ref. 30\% & AUC Ref. 0\% & AUC Ref. 50\% & AUC Ref. 30\% & AUC Ref. 0\% \\\hline
 
\multirow{5}{*}{EyePACS} & MAP & \uptriangle{OliveGreen}~97.4 $\pm$ 0.2 & \uptriangle{OliveGreen}~96.2 $\pm$ 0.2 & 95.0 $\pm$ 0.1 & \downtriangle{BrickRed}~86.5 $\pm$ 0.5 & \downtriangle{BrickRed}~87.5 $\pm$ 0.4 & \textbf{88.3} $\pm$ \textbf{0.2}\\
 & MC dropout & \uptriangle{OliveGreen}~\textbf{98.4} $\pm$ \textbf{0.1} & \uptriangle{OliveGreen}~\textbf{97.3} $\pm$ \textbf{0.2} & \textbf{95.7} $\pm$ \textbf{0.1} & \uptriangle{OliveGreen}~\textbf{90.4} $\pm$ \textbf{0.4} & \uptriangle{OliveGreen}~\textbf{89.5} $\pm$ \textbf{0.3} & 88.2 $\pm$ 0.2\\
 & MFVI & \uptriangle{OliveGreen}~97.6 $\pm$ 0.2 & \uptriangle{OliveGreen}~96.3 $\pm$ 0.2 & 94.8 $\pm$ 0.1 & \textcolor{Orange}{\textbf{=}}~87.7 $\pm$ 0.5 & \uptriangle{OliveGreen}~87.7 $\pm$ 0.3 & 86.2 $\pm$ 0.2\\
 & GVI & \uptriangle{OliveGreen}~98.2 $\pm$ 0.1 & \uptriangle{OliveGreen}~97.2 $\pm$ 0.1 & 94.9 $\pm$ 0.1 & \textcolor{Orange}{\textbf{=}}~87.7 $\pm$ 0.5 & \uptriangle{OliveGreen}~87.7 $\pm$ 0.4 & 87.0 $\pm$ 0.2\\
 & Radial & \uptriangle{OliveGreen}~97.8 $\pm$ 0.2 & \uptriangle{OliveGreen}~96.7 $\pm$ 0.2 & 94.8 $\pm$ 0.1 & \uptriangle{OliveGreen}~89.5 $\pm$ 0.4 & \uptriangle{OliveGreen}~88.9 $\pm$ 0.3 & 88.0 $\pm$ 0.2\\\hline
\multirow{5}{*}{KSSHP} & MAP & \uptriangle{OliveGreen}~96.4 $\pm$ 0.7 & \uptriangle{OliveGreen}~95.5 $\pm$ 0.7 & 93.9 $\pm$ 0.5 & \uptriangle{OliveGreen}~98.0 $\pm$ 0.8 & \uptriangle{OliveGreen}~97.0 $\pm$ 0.7 & \textbf{96.9} $\pm$ \textbf{0.3}\\
 & MC dropout & \uptriangle{OliveGreen}~97.3 $\pm$ 0.6 & \uptriangle{OliveGreen}~\textbf{96.7} $\pm$ \textbf{0.5} & \textbf{94.4} $\pm$ \textbf{0.4} & \uptriangle{OliveGreen}~\textbf{98.6} $\pm$ \textbf{0.5} & \uptriangle{OliveGreen}~\textbf{98.1} $\pm$ \textbf{0.5} & \textbf{96.9} $\pm$ \textbf{0.3}\\
 & MFVI & \uptriangle{OliveGreen}~96.8 $\pm$ 0.6 & \uptriangle{OliveGreen}~95.6 $\pm$ 0.6 & 93.4 $\pm$ 0.5 & \uptriangle{OliveGreen}~96.9 $\pm$ 0.8 & \downtriangle{BrickRed}~96.5 $\pm$ 0.7 & 96.6 $\pm$ 0.3\\
 & GVI & \uptriangle{OliveGreen}~\textbf{97.4} $\pm$ \textbf{0.4} & \uptriangle{OliveGreen}~95.9 $\pm$ 0.5 & 92.4 $\pm$ 0.5 & \uptriangle{OliveGreen}~97.4 $\pm$ 0.7 & \textcolor{Orange}{\textbf{=}}~96.5 $\pm$ 0.7 & 96.5 $\pm$ 0.3\\
 & Radial & \uptriangle{OliveGreen}~97.2 $\pm$ 0.5 & \uptriangle{OliveGreen}~95.8 $\pm$ 0.5 & 92.7 $\pm$ 0.5 & \uptriangle{OliveGreen}~97.5 $\pm$ 0.6 & \uptriangle{OliveGreen}~96.9 $\pm$ 0.6 & 96.4 $\pm$ 0.3\\\hline
\multirow{5}{*}{Messidor-2} & MAP & \uptriangle{OliveGreen}~99.1 $\pm$ 0.3 & \uptriangle{OliveGreen}~98.1 $\pm$ 0.4 & 95.9 $\pm$ 0.4 & \uptriangle{OliveGreen}~98.9 $\pm$ 0.4 & \uptriangle{OliveGreen}~97.7 $\pm$ 0.5 & 94.5 $\pm$ 0.5\\
 & MC dropout & \uptriangle{OliveGreen}~\textbf{99.7} $\pm$ \textbf{0.1} & \uptriangle{OliveGreen}~\textbf{99.1} $\pm$ \textbf{0.2} & \textbf{96.9} $\pm$ \textbf{0.4} & \uptriangle{OliveGreen}~99.2 $\pm$ 0.3 & \uptriangle{OliveGreen}~97.9 $\pm$ 0.4 & 94.8 $\pm$ 0.6\\
 & MFVI & \uptriangle{OliveGreen}~99.4 $\pm$ 0.3 & \uptriangle{OliveGreen}~98.8 $\pm$ 0.3 & 96.7 $\pm$ 0.4 & \uptriangle{OliveGreen}~97.9 $\pm$ 0.4 & \uptriangle{OliveGreen}~96.5 $\pm$ 0.5 & 92.4 $\pm$ 0.6\\
 & GVI & \uptriangle{OliveGreen}~99.0 $\pm$ 0.3 & \uptriangle{OliveGreen}~98.2 $\pm$ 0.4 & 96.0 $\pm$ 0.5 & \uptriangle{OliveGreen}~99.1 $\pm$ 0.4 & \uptriangle{OliveGreen}~98.2 $\pm$ 0.4 & 95.3 $\pm$ 0.5\\
 & Radial & \uptriangle{OliveGreen}~99.6 $\pm$ 0.2 & \uptriangle{OliveGreen}~98.6 $\pm$ 0.4 & 96.4 $\pm$ 0.5 & \uptriangle{OliveGreen}~\textbf{99.5} $\pm$ \textbf{0.3} & \uptriangle{OliveGreen}~\textbf{98.7} $\pm$ \textbf{0.3} & \textbf{95.8} $\pm$ \textbf{0.4}\\\hline
\multirow{5}{*}{APTOS} & MAP & \downtriangle{BrickRed}~\textbf{88.8} $\pm$ \textbf{1.0} & \downtriangle{BrickRed}~\textbf{93.6} $\pm$ \textbf{0.5} & 95.5 $\pm$ 0.3 & \downtriangle{BrickRed}~89.7 $\pm$ 0.8 & \uptriangle{OliveGreen}~91.0 $\pm$ 0.6 & 90.6 $\pm$ 0.5\\
 & MC dropout & \downtriangle{BrickRed}~86.2 $\pm$ 1.3 & \downtriangle{BrickRed}~92.7 $\pm$ 0.6 & 95.2 $\pm$ 0.4 & \uptriangle{OliveGreen}~80.7 $\pm$ 1.0 & \downtriangle{BrickRed}~79.9 $\pm$ 0.9 & 85.6 $\pm$ 0.7\\
 & MFVI & \downtriangle{BrickRed}~85.2 $\pm$ 1.3 & \downtriangle{BrickRed}~92.7 $\pm$ 0.6 & 95.6 $\pm$ 0.3 & \downtriangle{BrickRed}~71.0 $\pm$ 1.2 & \downtriangle{BrickRed}~75.2 $\pm$ 1.0 & 78.0 $\pm$ 0.8\\
 & GVI & \downtriangle{BrickRed}~82.3 $\pm$ 1.6 & \downtriangle{BrickRed}~91.6 $\pm$ 0.6 & 95.2 $\pm$ 0.4 & \downtriangle{BrickRed}~\textbf{89.8} $\pm$ \textbf{0.8} & \downtriangle{BrickRed}~90.6 $\pm$ 0.6 & 91.5 $\pm$ 0.5\\
 & Radial & \downtriangle{BrickRed}~84.5 $\pm$ 1.5 & \downtriangle{BrickRed}~92.9 $\pm$ 0.6 & \textbf{95.9} $\pm$ \textbf{0.3} & \downtriangle{BrickRed}~89.6 $\pm$ 0.8 & \downtriangle{BrickRed}~\textbf{92.4} $\pm$ \textbf{0.6} & \textbf{93.5} $\pm$ \textbf{0.4}\\\hline\hline
\multirow{5}{*}{EyePACS} & Deep ensemble & \uptriangle{OliveGreen}~97.9 $\pm$ 0.2 & \uptriangle{OliveGreen}~\textbf{97.1} $\pm$ \textbf{0.2} & \textbf{95.8} $\pm$ \textbf{0.1} & \downtriangle{BrickRed}~86.5 $\pm$ 0.6 & \downtriangle{BrickRed}~88.6 $\pm$ 0.3 & \textbf{89.1} $\pm$ \textbf{0.2}\\
 & MC dropout ensemble & \uptriangle{OliveGreen}~97.7 $\pm$ 0.2 & \uptriangle{OliveGreen}~96.7 $\pm$ 0.2 & 95.7 $\pm$ 0.1 & \uptriangle{OliveGreen}~91.2 $\pm$ 0.4 & \uptriangle{OliveGreen}~89.7 $\pm$ 0.3 & \textbf{89.1} $\pm$ \textbf{0.2}\\
 & MFVI ensemble & \uptriangle{OliveGreen}~\textbf{98.1} $\pm$ \textbf{0.2} & \uptriangle{OliveGreen}~96.9 $\pm$ 0.2 & 95.4 $\pm$ 0.1 & \uptriangle{OliveGreen}~\textbf{92.2} $\pm$ \textbf{0.3} & \uptriangle{OliveGreen}~\textbf{90.6} $\pm$ \textbf{0.3} & 88.2 $\pm$ 0.2\\
 & GVI ensemble & \uptriangle{OliveGreen}~97.9 $\pm$ 0.2 & \uptriangle{OliveGreen}~96.8 $\pm$ 0.2 & 95.5 $\pm$ 0.1 & \uptriangle{OliveGreen}~90.4 $\pm$ 0.4 & \uptriangle{OliveGreen}~89.8 $\pm$ 0.3 & 88.3 $\pm$ 0.2\\
 & Radial ensemble & \uptriangle{OliveGreen}~97.8 $\pm$ 0.2 & \uptriangle{OliveGreen}~96.6 $\pm$ 0.2 & 95.3 $\pm$ 0.1 & \downtriangle{BrickRed}~86.9 $\pm$ 0.5 & \downtriangle{BrickRed}~87.2 $\pm$ 0.4 & 88.6 $\pm$ 0.2\\\hline
\multirow{5}{*}{KSSHP} & Deep ensemble & \uptriangle{OliveGreen}~\textbf{98.0} $\pm$ \textbf{0.5} & \uptriangle{OliveGreen}~\textbf{97.0} $\pm$ \textbf{0.4} & \textbf{94.4} $\pm$ \textbf{0.4} & \uptriangle{OliveGreen}~98.1 $\pm$ 1.0 & \uptriangle{OliveGreen}~97.5 $\pm$ 0.7 & \textbf{97.4} $\pm$ \textbf{0.2}\\
 & MC dropout ensemble & \uptriangle{OliveGreen}~95.1 $\pm$ 0.9 & \uptriangle{OliveGreen}~94.3 $\pm$ 0.8 & 93.2 $\pm$ 0.5 & \uptriangle{OliveGreen}~\textbf{98.9} $\pm$ \textbf{0.5} & \uptriangle{OliveGreen}~\textbf{98.5} $\pm$ \textbf{0.5} & 97.3 $\pm$ 0.2\\
 & MFVI ensemble & \uptriangle{OliveGreen}~97.1 $\pm$ 0.6 & \uptriangle{OliveGreen}~95.6 $\pm$ 0.6 & 93.1 $\pm$ 0.5 & \uptriangle{OliveGreen}~98.1 $\pm$ 0.6 & \uptriangle{OliveGreen}~97.5 $\pm$ 0.5 & 96.8 $\pm$ 0.3\\
 & GVI ensemble & \uptriangle{OliveGreen}~97.7 $\pm$ 0.6 & \uptriangle{OliveGreen}~96.4 $\pm$ 0.6 & 93.8 $\pm$ 0.4 & \downtriangle{BrickRed}~96.8 $\pm$ 0.9 & \uptriangle{OliveGreen}~97.1 $\pm$ 0.7 & 96.9 $\pm$ 0.3\\
 & Radial ensemble & \uptriangle{OliveGreen}~97.2 $\pm$ 0.6 & \uptriangle{OliveGreen}~96.0 $\pm$ 0.6 & 93.4 $\pm$ 0.4 & \uptriangle{OliveGreen}~96.6 $\pm$ 0.8 & \downtriangle{BrickRed}~96.1 $\pm$ 0.7 & 96.7 $\pm$ 0.3\\\hline
\multirow{5}{*}{Messidor-2} & Deep ensemble & \uptriangle{OliveGreen}~99.6 $\pm$ 0.1 & \uptriangle{OliveGreen}~98.8 $\pm$ 0.3 & 96.8 $\pm$ 0.4 & \uptriangle{OliveGreen}~99.3 $\pm$ 0.4 & \uptriangle{OliveGreen}~98.1 $\pm$ 0.4 & 94.9 $\pm$ 0.6\\
 & MC dropout ensemble & \uptriangle{OliveGreen}~99.4 $\pm$ 0.2 & \uptriangle{OliveGreen}~98.7 $\pm$ 0.3 & 96.3 $\pm$ 0.4 & \uptriangle{OliveGreen}~98.6 $\pm$ 0.6 & \uptriangle{OliveGreen}~97.3 $\pm$ 0.6 & 94.3 $\pm$ 0.6\\
 & MFVI ensemble & \uptriangle{OliveGreen}~\textbf{99.9} $\pm$ \textbf{0.1} & \uptriangle{OliveGreen}~\textbf{99.5} $\pm$ \textbf{0.2} & \textbf{97.4} $\pm$ \textbf{0.3} & \uptriangle{OliveGreen}~99.9 $\pm$ 0.1 & \uptriangle{OliveGreen}~\textbf{99.6} $\pm$ \textbf{0.2} & 96.8 $\pm$ 0.4\\
 & GVI ensemble & \uptriangle{OliveGreen}~\textbf{99.9} $\pm$ \textbf{0.1} & \uptriangle{OliveGreen}~99.3 $\pm$ 0.2 & 97.2 $\pm$ 0.3 & \uptriangle{OliveGreen}~\textbf{100.0} $\pm$ \textbf{0.0} & \uptriangle{OliveGreen}~99.5 $\pm$ 0.2 & \textbf{97.0} $\pm$ \textbf{0.3}\\
 & Radial ensemble & \uptriangle{OliveGreen}~\textbf{99.9} $\pm$ \textbf{0.1} & \uptriangle{OliveGreen}~99.3 $\pm$ 0.2 & 97.1 $\pm$ 0.4 & \uptriangle{OliveGreen}~99.5 $\pm$ 0.3 & \uptriangle{OliveGreen}~99.0 $\pm$ 0.3 & 96.8 $\pm$ 0.4\\\hline
\multirow{5}{*}{APTOS} & Deep ensemble & \downtriangle{BrickRed}~88.9 $\pm$ 1.1 & \downtriangle{BrickRed}~93.1 $\pm$ 0.5 & 95.3 $\pm$ 0.4 & \downtriangle{BrickRed}~89.1 $\pm$ 0.8 & \downtriangle{BrickRed}~91.0 $\pm$ 0.6 & 92.2 $\pm$ 0.5\\
 & MC dropout ensemble & \downtriangle{BrickRed}~\textbf{90.6} $\pm$ \textbf{0.9} & \downtriangle{BrickRed}~93.6 $\pm$ 0.5 & 95.6 $\pm$ 0.3 & \downtriangle{BrickRed}~90.3 $\pm$ 0.8 & \downtriangle{BrickRed}~\textbf{92.9} $\pm$ \textbf{0.5} & \textbf{93.5} $\pm$ \textbf{0.4}\\
 & MFVI ensemble & \downtriangle{BrickRed}~86.9 $\pm$ 1.2 & \downtriangle{BrickRed}~93.5 $\pm$ 0.6 & \textbf{96.1} $\pm$ \textbf{0.3} & \downtriangle{BrickRed}~84.1 $\pm$ 1.2 & \downtriangle{BrickRed}~90.0 $\pm$ 0.7 & 93.4 $\pm$ 0.4\\
 & GVI ensemble & \downtriangle{BrickRed}~89.3 $\pm$ 0.9 & \downtriangle{BrickRed}~\textbf{93.8} $\pm$ \textbf{0.5} & 95.5 $\pm$ 0.3 & \downtriangle{BrickRed}~86.5 $\pm$ 1.0 & \downtriangle{BrickRed}~90.5 $\pm$ 0.7 & 92.7 $\pm$ 0.4\\
 & Radial ensemble & \downtriangle{BrickRed}~88.3 $\pm$ 1.1 & \downtriangle{BrickRed}~\textbf{93.8} $\pm$ \textbf{0.5} & 95.9 $\pm$ 0.3 & \downtriangle{BrickRed}~\textbf{91.1} $\pm$ \textbf{0.8} & \downtriangle{BrickRed}~\textbf{92.9} $\pm$ \textbf{0.6} & 93.3 $\pm$ 0.4\\

\end{tabular} 
} 
}
\caption{RDR results. Mean and standard deviation computed for 100 bootstrap resamples of the test data. Relative improvement to previous referral level is denoted with a green up-pointing triangle, orange equals sign, or red down-pointing triangle for increasing, equal, or worse performance, respectively.}
\label{tab:rdr_entropy}
\end{table}

\subsection{PIRC classification}
As no previous works have utilized robust neural networks for the 5-class clinical PIRC scheme, we can not directly compare the absolute performance. However, standard deep learning methods have been utilized, with similar data to our benchmark datasets, in \citet{Krause2018}. In the study, a deterministic Inception-v4 model was trained using over 1.6 million images from EyePACS affiliated clinics, 3 eye hospitals in India, one of them the same as the origin of the APTOS dataset, and the Messidor-2 dataset. The test set consisted of EyePACS originating images, on which the model achieved QWK of 84.0.

Visual illustration of PIRC results are presented in Figure \ref{fig:retained}B, full results in \ref{tab:pirc_entropy}. When no images are referred, the best performance is achieved with different ensembles: the deep ensemble achieves 81.3 QWK on the EyePACS dataset and 81.1 QWK on the KSSHP dataset. Similarly, the MFVI ensemble achieves 83.9 QWK on the Messidor-2 dataset, and the MC dropout ensemble achieves 87.9 QWK on the APTOS dataset. EyePACS trained models had better out-of-distribution performance than the KSSHP trained models.

While using entropy as the measure of uncertainty, the only EyePACS trained models that improved within distribution performance for all the referral levels were GVI, Radial, and GVI ensemble. These models also surpassed the 84.0 QWK when referring $\geq$ 30\% of images. Also, deep ensemble achieved 86.8 QWK and Radial ensemble 84.6 QWK for the 30\% referral level. All the models, except the MC dropout, consistently improve on the Messidor-2 dataset and reach over 84.0 QWK when referring $\geq$ 30\% of images. On the APTOS dataset, all models consistently improve for all referral levels and they surpass the 84.0 QWK when referring $\geq$ 30\% of images. However, we can see that on the KSSHP dataset, the models degrade consistently, apart from GVI and deep ensemble for 0\% to 30\% referral level, and no model reaches competitive performance on the KSSHP for any referral level.

When using the KSSHP data for training, we get significantly worse results compared to the EyePACS trained models, as no model consistently improves on the within distribution set, and overall, no model reaches the 84.0 QWK. Indeed, only the MFVI improves from 0\% to 30\% referral level. In terms of EyePACS generalization, all models improve from 0\% to 30\%, and Radial, MFVI ensemble, and Radial ensemble improve also from 30\% to 50\% referral level. However, no model reaches over 80.0 QWK. All models consistently improve for the Messidor-2 dataset and GVI, Radial, MC dropout ensemble, and Radial ensemble reach over 84.0 QWK for the 50\% referral level. However, the MFVI has generally poor performance, as the QWK only increases from 45.7 to 54.5. On the APTOS dataset, MC dropout, GVI, Radial, and all ensemble models improve from 0\% to 30\% referral level. On the 30\% to 50\% referral level, GVI and all ensembles except the GVI ensemble improve. Even though some models improve in the referral process, no model reaches over 80.0 QWK, the deep ensemble being closest with 79.9 QWK on the 50\% referral level.

The overall best performance for the EyePACS set is obtained using EyePACS trained Radial when referring 50\% of images, QWK of 87.9, and for the KSSHP set using KSSHP trained MFVI and referring 30\% of images, QWK of 81.6. For the out-of-distribution sets, the best Messidor-2 results are obtained using EyePACS trained Radial when referring 50\% of images, QWK of 94.3, and for the APTOS set using EyePACS trained MFVI ensemble when referring 50\% of images, QWK of 97.3. Since the uncertainty estimation is motivated from the point of view of clinical interest, it is extremely concerning that the methods appear to not work in a similar manner when trained on a clinical dataset in comparison to the benchmark datasets, especially on the clinical set itself.

\begin{table}[h]
\centering
\makebox[\textwidth]{
\resizebox{\textwidth}{!}{%
\begin{tabular}{c|c|l|l|l||l|l|l}
 \multirow{2}{*}{Test Dataset} & \multirow{2}{*}{Method} & \multicolumn{3}{c||}{EyePACS trained} & \multicolumn{3}{c}{KSSHP trained} \\
 & & QWK Ref. 50\% & QWK Ref. 30\% & QWK Ref. 0\% & QWK Ref. 50\% & QWK Ref. 30\% & QWK Ref. 0\% \\\hline

\multirow{5}{*}{EyePACS} & MAP & \downtriangle{BrickRed}~81.3 $\pm$ 1.0 & \uptriangle{OliveGreen}~82.0 $\pm$ 0.6 & 79.7 $\pm$ 0.4 & \downtriangle{BrickRed}~57.4 $\pm$ 1.0 & \uptriangle{OliveGreen}~63.3 $\pm$ 0.7 & 62.7 $\pm$ 0.4\\
 & MC dropout & \downtriangle{BrickRed}~79.5 $\pm$ 1.2 & \uptriangle{OliveGreen}~80.0 $\pm$ 0.7 & 79.9 $\pm$ 0.4 & \downtriangle{BrickRed}~67.6 $\pm$ 0.8 & \uptriangle{OliveGreen}~69.3 $\pm$ 0.5 & 62.5 $\pm$ 0.4\\
 & MFVI & \downtriangle{BrickRed}~79.0 $\pm$ 1.1 & \uptriangle{OliveGreen}~80.5 $\pm$ 0.7 & 78.4 $\pm$ 0.4 & \downtriangle{BrickRed}~67.7 $\pm$ 0.7 & \uptriangle{OliveGreen}~69.3 $\pm$ 0.5 & 60.2 $\pm$ 0.5\\
 & GVI & \uptriangle{OliveGreen}~86.9 $\pm$ 0.6 & \uptriangle{OliveGreen}~86.2 $\pm$ 0.4 & 79.0 $\pm$ 0.3 & \downtriangle{BrickRed}~68.8 $\pm$ 0.9 & \uptriangle{OliveGreen}~70.9 $\pm$ 0.6 & \textbf{64.1} $\pm$ \textbf{0.5}\\
 & Radial & \uptriangle{OliveGreen}~\textbf{87.9} $\pm$ \textbf{0.5} & \uptriangle{OliveGreen}~\textbf{87.0} $\pm$ \textbf{0.4} & \textbf{80.2} $\pm$ \textbf{0.3} & \uptriangle{OliveGreen}~\textbf{73.8} $\pm$ \textbf{0.8} & \uptriangle{OliveGreen}~\textbf{72.8} $\pm$ \textbf{0.5} & 62.0 $\pm$ 0.4\\\hline
\multirow{5}{*}{KSSHP} & MAP & \downtriangle{BrickRed}~47.6 $\pm$ 3.1 & \downtriangle{BrickRed}~58.8 $\pm$ 2.0 & 67.5 $\pm$ 0.9 & \downtriangle{BrickRed}~70.9 $\pm$ 1.6 & \downtriangle{BrickRed}~76.0 $\pm$ 1.0 & \textbf{81.0} $\pm$ \textbf{0.6}\\
 & MC dropout & \downtriangle{BrickRed}~32.6 $\pm$ 4.8 & \downtriangle{BrickRed}~47.2 $\pm$ 2.7 & \textbf{67.7} $\pm$ \textbf{0.8} & \downtriangle{BrickRed}~72.1 $\pm$ 1.4 & \downtriangle{BrickRed}~75.3 $\pm$ 1.1 & 80.3 $\pm$ 0.6\\
 & MFVI & \downtriangle{BrickRed}~40.5 $\pm$ 3.2 & \downtriangle{BrickRed}~45.3 $\pm$ 2.4 & 63.8 $\pm$ 0.9 & \downtriangle{BrickRed}~\textbf{81.4} $\pm$ \textbf{1.1} & \uptriangle{OliveGreen}~\textbf{81.6} $\pm$ \textbf{0.8} & 79.9 $\pm$ 0.6\\
 & GVI & \downtriangle{BrickRed}~\textbf{61.1} $\pm$ \textbf{1.8} & \uptriangle{OliveGreen}~\textbf{65.6} $\pm$ \textbf{1.2} & 62.8 $\pm$ 0.8 & \downtriangle{BrickRed}~74.8 $\pm$ 1.7 & \downtriangle{BrickRed}~76.1 $\pm$ 1.1 & 80.0 $\pm$ 0.6\\
 & Radial & \downtriangle{BrickRed}~49.9 $\pm$ 2.9 & \downtriangle{BrickRed}~59.6 $\pm$ 2.1 & 65.4 $\pm$ 0.9 & \downtriangle{BrickRed}~77.6 $\pm$ 1.6 & \downtriangle{BrickRed}~78.5 $\pm$ 1.3 & 79.9 $\pm$ 0.6\\\hline
\multirow{5}{*}{Messidor-2} & MAP & \uptriangle{OliveGreen}~91.7 $\pm$ 1.3 & \uptriangle{OliveGreen}~\textbf{90.0} $\pm$ \textbf{1.0} & 80.7 $\pm$ 1.2 & \uptriangle{OliveGreen}~83.2 $\pm$ 1.2 & \uptriangle{OliveGreen}~77.6 $\pm$ 1.3 & 66.7 $\pm$ 1.4\\
 & MC dropout & \downtriangle{BrickRed}~88.5 $\pm$ 1.8 & \uptriangle{OliveGreen}~89.5 $\pm$ 1.2 & \textbf{81.8} $\pm$ \textbf{1.4} & \uptriangle{OliveGreen}~82.5 $\pm$ 1.3 & \uptriangle{OliveGreen}~77.2 $\pm$ 1.3 & 64.3 $\pm$ 1.4\\
 & MFVI & \uptriangle{OliveGreen}~90.7 $\pm$ 1.3 & \uptriangle{OliveGreen}~89.3 $\pm$ 1.2 & 75.3 $\pm$ 1.6 & \uptriangle{OliveGreen}~54.5 $\pm$ 2.3 & \uptriangle{OliveGreen}~51.8 $\pm$ 1.7 & 45.7 $\pm$ 1.7\\
 & GVI & \uptriangle{OliveGreen}~91.8 $\pm$ 1.1 & \uptriangle{OliveGreen}~87.2 $\pm$ 1.1 & 73.3 $\pm$ 1.4 & \uptriangle{OliveGreen}~85.8 $\pm$ 1.2 & \uptriangle{OliveGreen}~80.2 $\pm$ 1.1 & 68.1 $\pm$ 1.4\\
 & Radial & \uptriangle{OliveGreen}~\textbf{94.3} $\pm$ \textbf{0.9} & \uptriangle{OliveGreen}~88.8 $\pm$ 1.1 & 76.8 $\pm$ 1.3 & \uptriangle{OliveGreen}~\textbf{86.6} $\pm$ \textbf{0.8} & \uptriangle{OliveGreen}~\textbf{81.9} $\pm$ \textbf{1.0} & \textbf{69.8} $\pm$ \textbf{1.3}\\\hline
\multirow{5}{*}{APTOS} & MAP & \uptriangle{OliveGreen}~94.8 $\pm$ 0.6 & \uptriangle{OliveGreen}~90.5 $\pm$ 0.6 & 83.1 $\pm$ 0.6 & \downtriangle{BrickRed}~53.6 $\pm$ 1.4 & \downtriangle{BrickRed}~56.3 $\pm$ 1.0 & 56.8 $\pm$ 0.9\\
 & MC dropout & \uptriangle{OliveGreen}~\textbf{95.9} $\pm$ \textbf{0.5} & \uptriangle{OliveGreen}~91.2 $\pm$ 0.5 & 83.0 $\pm$ 0.6 & \downtriangle{BrickRed}~49.1 $\pm$ 1.3 & \uptriangle{OliveGreen}~51.5 $\pm$ 1.0 & 50.4 $\pm$ 0.9\\
 & MFVI & \uptriangle{OliveGreen}~94.5 $\pm$ 0.7 & \uptriangle{OliveGreen}~89.8 $\pm$ 0.6 & 80.4 $\pm$ 0.6 & \downtriangle{BrickRed}~33.7 $\pm$ 1.3 & \downtriangle{BrickRed}~38.8 $\pm$ 1.2 & 41.3 $\pm$ 1.0\\
 & GVI & \uptriangle{OliveGreen}~94.6 $\pm$ 0.5 & \uptriangle{OliveGreen}~\textbf{91.4} $\pm$ \textbf{0.4} & 84.6 $\pm$ 0.5 & \uptriangle{OliveGreen}~73.4 $\pm$ 1.1 & \uptriangle{OliveGreen}~69.6 $\pm$ 1.0 & 59.0 $\pm$ 1.0\\
 & Radial & \uptriangle{OliveGreen}~92.7 $\pm$ 0.4 & \uptriangle{OliveGreen}~90.1 $\pm$ 0.5 & \textbf{85.2} $\pm$ \textbf{0.5} & \downtriangle{BrickRed}~\textbf{77.1} $\pm$ \textbf{1.0} & \uptriangle{OliveGreen}~\textbf{77.7} $\pm$ \textbf{0.8} & \textbf{73.0} $\pm$ \textbf{0.7}\\\hline\hline
\multirow{5}{*}{EyePACS} & Deep ensemble & \downtriangle{BrickRed}~\textbf{86.4} $\pm$ \textbf{0.6} & \uptriangle{OliveGreen}~\textbf{86.8} $\pm$ \textbf{0.4} & \textbf{81.3} $\pm$ \textbf{0.4} & \downtriangle{BrickRed}~66.4 $\pm$ 0.9 & \uptriangle{OliveGreen}~70.3 $\pm$ 0.6 & \textbf{64.5} $\pm$ \textbf{0.4}\\
 & MC dropout ensemble & \downtriangle{BrickRed}~80.1 $\pm$ 1.1 & \uptriangle{OliveGreen}~80.9 $\pm$ 0.7 & 79.2 $\pm$ 0.4 & \downtriangle{BrickRed}~64.6 $\pm$ 1.0 & \uptriangle{OliveGreen}~71.3 $\pm$ 0.5 & 64.3 $\pm$ 0.4\\
 & MFVI ensemble & \uptriangle{OliveGreen}~77.9 $\pm$ 1.6 & \downtriangle{BrickRed}~77.5 $\pm$ 1.0 & 80.6 $\pm$ 0.3 & \uptriangle{OliveGreen}~73.4 $\pm$ 0.6 & \uptriangle{OliveGreen}~72.9 $\pm$ 0.4 & 62.5 $\pm$ 0.4\\
 & GVI ensemble & \uptriangle{OliveGreen}~84.6 $\pm$ 0.8 & \uptriangle{OliveGreen}~84.3 $\pm$ 0.6 & 80.7 $\pm$ 0.4 & \downtriangle{BrickRed}~70.1 $\pm$ 0.8 & \uptriangle{OliveGreen}~\textbf{73.7} $\pm$ \textbf{0.5} & 63.1 $\pm$ 0.4\\
 & Radial ensemble & \downtriangle{BrickRed}~83.9 $\pm$ 0.9 & \uptriangle{OliveGreen}~84.6 $\pm$ 0.5 & 80.7 $\pm$ 0.3 & \uptriangle{OliveGreen}~\textbf{74.2} $\pm$ \textbf{0.6} & \uptriangle{OliveGreen}~73.3 $\pm$ 0.5 & 61.6 $\pm$ 0.4\\\hline
\multirow{5}{*}{KSSHP} & Deep ensemble & \downtriangle{BrickRed}~\textbf{62.3} $\pm$ \textbf{2.0} & \uptriangle{OliveGreen}~\textbf{73.1} $\pm$ \textbf{1.1} & \textbf{69.6} $\pm$ \textbf{0.8} & \downtriangle{BrickRed}~\textbf{78.4} $\pm$ \textbf{1.3} & \downtriangle{BrickRed}~\textbf{78.8} $\pm$ \textbf{1.0} & \textbf{81.1} $\pm$ \textbf{0.5}\\
 & MC dropout ensemble & \downtriangle{BrickRed}~43.7 $\pm$ 3.9 & \downtriangle{BrickRed}~56.0 $\pm$ 2.0 & 68.4 $\pm$ 0.9 & \downtriangle{BrickRed}~71.3 $\pm$ 1.8 & \downtriangle{BrickRed}~75.0 $\pm$ 1.0 & 80.6 $\pm$ 0.6\\
 & MFVI ensemble & \downtriangle{BrickRed}~24.4 $\pm$ 5.3 & \downtriangle{BrickRed}~39.4 $\pm$ 2.9 & 66.1 $\pm$ 0.9 & \downtriangle{BrickRed}~74.7 $\pm$ 1.5 & \downtriangle{BrickRed}~77.7 $\pm$ 1.1 & 80.5 $\pm$ 0.6\\
 & GVI ensemble & \downtriangle{BrickRed}~40.0 $\pm$ 3.4 & \downtriangle{BrickRed}~55.0 $\pm$ 1.9 & 66.8 $\pm$ 0.9 & \downtriangle{BrickRed}~66.6 $\pm$ 2.0 & \downtriangle{BrickRed}~76.4 $\pm$ 1.1 & 80.9 $\pm$ 0.5\\
 & Radial ensemble & \downtriangle{BrickRed}~33.4 $\pm$ 3.5 & \downtriangle{BrickRed}~56.2 $\pm$ 2.2 & 66.2 $\pm$ 1.0 & \downtriangle{BrickRed}~73.2 $\pm$ 1.6 & \downtriangle{BrickRed}~77.8 $\pm$ 1.1 & 80.2 $\pm$ 0.6\\\hline
\multirow{5}{*}{Messidor-2} & Deep ensemble & \uptriangle{OliveGreen}~\textbf{93.8} $\pm$ \textbf{1.0} & \uptriangle{OliveGreen}~91.1 $\pm$ 0.9 & 81.4 $\pm$ 1.2 & \uptriangle{OliveGreen}~83.2 $\pm$ 1.2 & \uptriangle{OliveGreen}~77.6 $\pm$ 1.2 & 63.7 $\pm$ 1.5\\
 & MC dropout ensemble & \uptriangle{OliveGreen}~93.5 $\pm$ 1.1 & \uptriangle{OliveGreen}~\textbf{92.1} $\pm$ \textbf{0.8} & 80.3 $\pm$ 1.3 & \uptriangle{OliveGreen}~\textbf{88.1} $\pm$ \textbf{1.1} & \uptriangle{OliveGreen}~\textbf{80.4} $\pm$ \textbf{1.3} & \textbf{68.0} $\pm$ \textbf{1.5}\\
 & MFVI ensemble & \uptriangle{OliveGreen}~93.2 $\pm$ 1.3 & \uptriangle{OliveGreen}~91.3 $\pm$ 1.0 & \textbf{83.9} $\pm$ \textbf{1.2} & \uptriangle{OliveGreen}~83.2 $\pm$ 1.4 & \uptriangle{OliveGreen}~73.2 $\pm$ 1.6 & 59.9 $\pm$ 1.5\\
 & GVI ensemble & \uptriangle{OliveGreen}~92.6 $\pm$ 1.2 & \uptriangle{OliveGreen}~89.9 $\pm$ 1.0 & 80.0 $\pm$ 1.4 & \uptriangle{OliveGreen}~82.6 $\pm$ 1.5 & \uptriangle{OliveGreen}~72.3 $\pm$ 1.5 & 58.4 $\pm$ 1.4\\
 & Radial ensemble & \uptriangle{OliveGreen}~93.7 $\pm$ 1.0 & \uptriangle{OliveGreen}~91.0 $\pm$ 1.0 & 82.2 $\pm$ 1.3 & \uptriangle{OliveGreen}~84.4 $\pm$ 1.2 & \uptriangle{OliveGreen}~77.2 $\pm$ 1.2 & 63.1 $\pm$ 1.5\\\hline
\multirow{5}{*}{APTOS} & Deep ensemble & \uptriangle{OliveGreen}~94.8 $\pm$ 0.5 & \uptriangle{OliveGreen}~92.2 $\pm$ 0.5 & 85.4 $\pm$ 0.5 & \uptriangle{OliveGreen}~\textbf{79.9} $\pm$ \textbf{0.8} & \uptriangle{OliveGreen}~\textbf{77.6} $\pm$ \textbf{0.8} & \textbf{71.0} $\pm$ \textbf{0.7}\\
 & MC dropout ensemble & \uptriangle{OliveGreen}~97.0 $\pm$ 0.4 & \uptriangle{OliveGreen}~\textbf{93.7} $\pm$ \textbf{0.4} & \textbf{87.9} $\pm$ \textbf{0.4} & \uptriangle{OliveGreen}~60.8 $\pm$ 1.3 & \uptriangle{OliveGreen}~58.6 $\pm$ 1.1 & 53.0 $\pm$ 1.0\\
 & MFVI ensemble & \uptriangle{OliveGreen}~\textbf{97.3} $\pm$ \textbf{0.4} & \uptriangle{OliveGreen}~93.1 $\pm$ 0.4 & 84.3 $\pm$ 0.6 & \uptriangle{OliveGreen}~70.2 $\pm$ 1.0 & \uptriangle{OliveGreen}~69.9 $\pm$ 0.8 & 60.7 $\pm$ 0.9\\
 & GVI ensemble & \uptriangle{OliveGreen}~96.7 $\pm$ 0.4 & \uptriangle{OliveGreen}~93.5 $\pm$ 0.4 & 87.3 $\pm$ 0.5 & \downtriangle{BrickRed}~70.3 $\pm$ 1.1 & \uptriangle{OliveGreen}~71.8 $\pm$ 0.9 & 59.2 $\pm$ 0.8\\
 & Radial ensemble & \uptriangle{OliveGreen}~94.8 $\pm$ 0.4 & \uptriangle{OliveGreen}~92.2 $\pm$ 0.4 & 87.1 $\pm$ 0.4 & \uptriangle{OliveGreen}~68.9 $\pm$ 1.2 & \uptriangle{OliveGreen}~68.6 $\pm$ 1.0 & 60.6 $\pm$ 0.9\\

\end{tabular} 
} 
}
\caption{PIRC results using posterior predictive entropy as the uncertainty measure. Mean and standard deviation computed for 100 bootstrap resamples of the test data. Relative improvement to previous referral level is denoted with a green up-pointing triangle, orange equals sign, or red down-pointing triangle for increasing, equal, or worse performance, respectively.}
\label{tab:pirc_entropy}
\end{table}

\subsection{QWK-Risk as alternative uncertainty measure}

Our proposed QWK-Risk uncertainty measure results are presented in Figure \ref{fig:retained}C and Table \ref{tab:pirc_qwkrisk}. We can see that the QWK-Risk uncertainty based method systematically improves both the within distribution test results and cross out-of-distribution results for the two train sets. Within distribution for the EyePACS set, all models reach over 84.0 QWK when referring $\geq$ 30\% of images. No EyePACS trained model reaches the 84.0 QWK on the KSSHP set. However, QWK-Risk enables deep ensemble to reach $\geq$ 80.0 QWK for all referral levels. In addition, deterministic MAP network, MC dropout ensemble, and MFVI ensemble reach $\geq$ 80.0 QWK for 50\% referral level. On the Messidor-2 set, the performance of some models decreases and some improve in comparison to the entropy based uncertainty, most notably the GVI no longer reaches over 84.0 QWK but all other models reach over 84.0 QWK for referral levels $\geq$ 30\%. Systematic decrease in the performance is seen for the APTOS dataset, and now only MC dropout ensemble and GVI ensemble reach over 84.0 QWK when referring 30\% of images.

From the clinical perspective, an important finding is that using the QWK-Risk, all our models reach $\geq 84.0$ QWK for referral levels $\geq$ 30\% when trained and tested on the KSSHP set. Indeed, the GVI ensemble even reaches 89.9 QWK for the within KSSHP distribution test when 50\% of examples are referred. The generalization of the uncertainty estimates to the EyePACS set also increases and now the deep ensemble, MC dropout ensemble, and MFVI ensemble reach $\geq 80.0$ QWK, however, no model can reach the 84.0 QWK. The QWK-Risk decreases the performance for the Messidor-2 and APTOS datasets in comparison to entropy.

Using the QWK-Risk as the uncertainty measure, the best performance for the EyePACS set is obtained using EyePACS trained MC dropout ensemble and Radial ensemble when referring 50\% of images, QWK of 92.6, for the KSSHP set using KSSHP trained GVI ensemble and referring 50\% of images, QWK of 89.9, for the Messidor-2 using EyePACS trained MC dropout ensemble and MFVI ensemble when referring 50\% of images, QWK of 94.3, and for the APTOS set using EyePACS trained MC dropout ensemble when referring 0\% of images, QWK of 87.9.

\begin{table}[h]
\centering
\makebox[\textwidth]{
\resizebox{\textwidth}{!}{%
\begin{tabular}{c|c|l|l|l||l|l|l}
 \multirow{2}{*}{Test Dataset} & \multirow{2}{*}{Method} & \multicolumn{3}{c||}{EyePACS trained} & \multicolumn{3}{c}{KSSHP trained} \\
 & & QWK Ref. 50\% & QWK Ref. 30\% & QWK Ref. 0\% & QWK Ref. 50\% & QWK Ref. 30\% & QWK Ref. 0\% \\\hline

\multirow{5}{*}{EyePACS} & MAP & \uptriangle{OliveGreen}~\textbf{92.1} $\pm$ \textbf{0.2} & \uptriangle{OliveGreen}~90.3 $\pm$ 0.2 & 79.7 $\pm$ 0.4 & \uptriangle{OliveGreen}~77.8 $\pm$ 0.4 & \uptriangle{OliveGreen}~73.7 $\pm$ 0.4 & 62.7 $\pm$ 0.4\\
 & MC dropout & \uptriangle{OliveGreen}~92.0 $\pm$ 0.2 & \uptriangle{OliveGreen}~\textbf{90.6} $\pm$ \textbf{0.2} & 79.9 $\pm$ 0.4 & \uptriangle{OliveGreen}~78.5 $\pm$ 0.3 & \uptriangle{OliveGreen}~75.1 $\pm$ 0.3 & 62.5 $\pm$ 0.4\\
 & MFVI & \uptriangle{OliveGreen}~91.0 $\pm$ 0.2 & \uptriangle{OliveGreen}~89.5 $\pm$ 0.2 & 78.4 $\pm$ 0.4 & \uptriangle{OliveGreen}~71.5 $\pm$ 0.4 & \uptriangle{OliveGreen}~69.9 $\pm$ 0.4 & 60.2 $\pm$ 0.5\\
 & GVI & \uptriangle{OliveGreen}~89.4 $\pm$ 0.2 & \uptriangle{OliveGreen}~88.2 $\pm$ 0.3 & 79.0 $\pm$ 0.3 & \uptriangle{OliveGreen}~\textbf{79.9} $\pm$ \textbf{0.4} & \uptriangle{OliveGreen}~\textbf{76.6} $\pm$ \textbf{0.4} & \textbf{64.1} $\pm$ \textbf{0.5}\\
 & Radial & \uptriangle{OliveGreen}~91.3 $\pm$ 0.2 & \uptriangle{OliveGreen}~89.9 $\pm$ 0.2 & \textbf{80.2} $\pm$ \textbf{0.3} & \uptriangle{OliveGreen}~78.2 $\pm$ 0.4 & \uptriangle{OliveGreen}~74.6 $\pm$ 0.4 & 62.0 $\pm$ 0.4\\\hline
\multirow{5}{*}{KSSHP} & MAP & \uptriangle{OliveGreen}~\textbf{81.4} $\pm$ \textbf{0.9} & \uptriangle{OliveGreen}~\textbf{78.6} $\pm$ \textbf{0.9} & 67.5 $\pm$ 0.9 & \uptriangle{OliveGreen}~\textbf{89.4} $\pm$ \textbf{0.5} & \uptriangle{OliveGreen}~\textbf{87.9} $\pm$ \textbf{0.5} & \textbf{81.0} $\pm$ \textbf{0.6}\\
 & MC dropout & \uptriangle{OliveGreen}~79.6 $\pm$ 1.0 & \uptriangle{OliveGreen}~77.5 $\pm$ 0.9 & \textbf{67.7} $\pm$ \textbf{0.8} & \uptriangle{OliveGreen}~88.6 $\pm$ 0.5 & \uptriangle{OliveGreen}~87.1 $\pm$ 0.5 & 80.3 $\pm$ 0.6\\
 & MFVI & \uptriangle{OliveGreen}~76.1 $\pm$ 1.0 & \uptriangle{OliveGreen}~74.5 $\pm$ 0.9 & 63.8 $\pm$ 0.9 & \uptriangle{OliveGreen}~87.3 $\pm$ 0.5 & \uptriangle{OliveGreen}~85.9 $\pm$ 0.5 & 79.9 $\pm$ 0.6\\
 & GVI & \uptriangle{OliveGreen}~75.8 $\pm$ 0.8 & \uptriangle{OliveGreen}~73.6 $\pm$ 0.8 & 62.8 $\pm$ 0.8 & \uptriangle{OliveGreen}~87.9 $\pm$ 0.4 & \uptriangle{OliveGreen}~86.5 $\pm$ 0.4 & 80.0 $\pm$ 0.6\\
 & Radial & \uptriangle{OliveGreen}~78.7 $\pm$ 0.9 & \uptriangle{OliveGreen}~76.6 $\pm$ 0.8 & 65.4 $\pm$ 0.9 & \uptriangle{OliveGreen}~87.8 $\pm$ 0.5 & \uptriangle{OliveGreen}~86.5 $\pm$ 0.5 & 79.9 $\pm$ 0.6\\\hline
\multirow{5}{*}{Messidor-2} & MAP & \uptriangle{OliveGreen}~91.9 $\pm$ 0.8 & \uptriangle{OliveGreen}~89.9 $\pm$ 0.8 & 80.7 $\pm$ 1.2 & \uptriangle{OliveGreen}~\textbf{74.8} $\pm$ \textbf{1.4} & \uptriangle{OliveGreen}~73.4 $\pm$ 1.3 & 66.7 $\pm$ 1.4\\
 & MC dropout & \uptriangle{OliveGreen}~\textbf{94.0} $\pm$ \textbf{0.6} & \uptriangle{OliveGreen}~\textbf{91.7} $\pm$ \textbf{0.7} & \textbf{81.8} $\pm$ \textbf{1.4} & \downtriangle{BrickRed}~63.1 $\pm$ 2.1 & \uptriangle{OliveGreen}~67.9 $\pm$ 1.5 & 64.3 $\pm$ 1.4\\
 & MFVI & \uptriangle{OliveGreen}~86.3 $\pm$ 1.6 & \uptriangle{OliveGreen}~84.8 $\pm$ 1.4 & 75.3 $\pm$ 1.6 & \downtriangle{BrickRed}~18.3 $\pm$ 1.3 & \downtriangle{BrickRed}~35.4 $\pm$ 1.8 & 45.7 $\pm$ 1.7\\
 & GVI & \uptriangle{OliveGreen}~83.0 $\pm$ 1.5 & \uptriangle{OliveGreen}~81.7 $\pm$ 1.3 & 73.3 $\pm$ 1.4 & \downtriangle{BrickRed}~70.5 $\pm$ 1.8 & \uptriangle{OliveGreen}~73.4 $\pm$ 1.3 & 68.1 $\pm$ 1.4\\
 & Radial & \uptriangle{OliveGreen}~90.0 $\pm$ 1.0 & \uptriangle{OliveGreen}~87.4 $\pm$ 1.0 & 76.8 $\pm$ 1.3 & \downtriangle{BrickRed}~73.5 $\pm$ 1.4 & \uptriangle{OliveGreen}~\textbf{74.6} $\pm$ \textbf{1.1} & \textbf{69.8} $\pm$ \textbf{1.3}\\\hline
\multirow{5}{*}{APTOS} & MAP & \downtriangle{BrickRed}~\textbf{66.5} $\pm$ \textbf{2.4} & \downtriangle{BrickRed}~\textbf{82.8} $\pm$ \textbf{0.7} & 83.1 $\pm$ 0.6 & \downtriangle{BrickRed}~19.7 $\pm$ 1.3 & \downtriangle{BrickRed}~47.3 $\pm$ 1.5 & 56.8 $\pm$ 0.9\\
 & MC dropout & \downtriangle{BrickRed}~51.4 $\pm$ 4.7 & \downtriangle{BrickRed}~79.0 $\pm$ 0.9 & 83.0 $\pm$ 0.6 & \downtriangle{BrickRed}~16.6 $\pm$ 1.1 & \downtriangle{BrickRed}~41.6 $\pm$ 1.3 & 50.4 $\pm$ 0.9\\
 & MFVI & \downtriangle{BrickRed}~50.8 $\pm$ 4.1 & \downtriangle{BrickRed}~77.9 $\pm$ 0.8 & 80.4 $\pm$ 0.6 & \uptriangle{OliveGreen}~18.7 $\pm$ 1.2 & \downtriangle{BrickRed}~17.2 $\pm$ 1.1 & 41.3 $\pm$ 1.0\\
 & GVI & \downtriangle{BrickRed}~45.1 $\pm$ 4.4 & \downtriangle{BrickRed}~80.4 $\pm$ 0.8 & 84.6 $\pm$ 0.5 & \downtriangle{BrickRed}~22.0 $\pm$ 5.4 & \downtriangle{BrickRed}~57.9 $\pm$ 1.4 & 59.0 $\pm$ 1.0\\
 & Radial & \downtriangle{BrickRed}~52.5 $\pm$ 5.0 & \downtriangle{BrickRed}~81.3 $\pm$ 0.8 & \textbf{85.2} $\pm$ \textbf{0.5} & \downtriangle{BrickRed}~\textbf{35.8} $\pm$ \textbf{5.9} & \downtriangle{BrickRed}~\textbf{69.5} $\pm$ \textbf{1.1} & \textbf{73.0} $\pm$ \textbf{0.7}\\\hline\hline
\multirow{5}{*}{EyePACS} & Deep ensemble & \uptriangle{OliveGreen}~92.2 $\pm$ 0.2 & \uptriangle{OliveGreen}~90.9 $\pm$ 0.2 & \textbf{81.3} $\pm$ \textbf{0.4} & \uptriangle{OliveGreen}~80.5 $\pm$ 0.3 & \uptriangle{OliveGreen}~\textbf{77.4} $\pm$ \textbf{0.4} & \textbf{64.5} $\pm$ \textbf{0.4}\\
 & MC dropout ensemble & \uptriangle{OliveGreen}~\textbf{92.6} $\pm$ \textbf{0.2} & \uptriangle{OliveGreen}~90.7 $\pm$ 0.2 & 79.2 $\pm$ 0.4 & \uptriangle{OliveGreen}~\textbf{81.3} $\pm$ \textbf{0.3} & \uptriangle{OliveGreen}~\textbf{77.4} $\pm$ \textbf{0.3} & 64.3 $\pm$ 0.4\\
 & MFVI ensemble & \uptriangle{OliveGreen}~92.5 $\pm$ 0.2 & \uptriangle{OliveGreen}~90.8 $\pm$ 0.2 & 80.6 $\pm$ 0.3 & \uptriangle{OliveGreen}~80.0 $\pm$ 0.3 & \uptriangle{OliveGreen}~75.9 $\pm$ 0.3 & 62.5 $\pm$ 0.4\\
 & GVI ensemble & \uptriangle{OliveGreen}~92.3 $\pm$ 0.2 & \uptriangle{OliveGreen}~90.9 $\pm$ 0.2 & 80.7 $\pm$ 0.4 & \uptriangle{OliveGreen}~79.3 $\pm$ 0.3 & \uptriangle{OliveGreen}~75.7 $\pm$ 0.4 & 63.1 $\pm$ 0.4\\
 & Radial ensemble & \uptriangle{OliveGreen}~\textbf{92.6} $\pm$ \textbf{0.2} & \uptriangle{OliveGreen}~\textbf{91.0} $\pm$ \textbf{0.2} & 80.7 $\pm$ 0.3 & \uptriangle{OliveGreen}~78.7 $\pm$ 0.3 & \uptriangle{OliveGreen}~74.7 $\pm$ 0.4 & 61.6 $\pm$ 0.4\\\hline
\multirow{5}{*}{KSSHP} & Deep ensemble & \uptriangle{OliveGreen}~\textbf{81.6} $\pm$ \textbf{0.8} & \uptriangle{OliveGreen}~\textbf{80.2} $\pm$ \textbf{0.8} & \textbf{69.6} $\pm$ \textbf{0.8} & \uptriangle{OliveGreen}~89.5 $\pm$ 0.4 & \uptriangle{OliveGreen}~88.1 $\pm$ 0.4 & \textbf{81.1} $\pm$ \textbf{0.5}\\
 & MC dropout ensemble & \uptriangle{OliveGreen}~81.1 $\pm$ 0.9 & \uptriangle{OliveGreen}~78.9 $\pm$ 0.9 & 68.4 $\pm$ 0.9 & \uptriangle{OliveGreen}~89.4 $\pm$ 0.4 & \uptriangle{OliveGreen}~87.9 $\pm$ 0.5 & 80.6 $\pm$ 0.6\\
 & MFVI ensemble & \uptriangle{OliveGreen}~80.0 $\pm$ 0.9 & \uptriangle{OliveGreen}~77.9 $\pm$ 0.9 & 66.1 $\pm$ 0.9 & \uptriangle{OliveGreen}~89.7 $\pm$ 0.4 & \uptriangle{OliveGreen}~87.7 $\pm$ 0.4 & 80.5 $\pm$ 0.6\\
 & GVI ensemble & \uptriangle{OliveGreen}~79.3 $\pm$ 0.9 & \uptriangle{OliveGreen}~76.7 $\pm$ 0.9 & 66.8 $\pm$ 0.9 & \uptriangle{OliveGreen}~\textbf{89.9} $\pm$ \textbf{0.4} & \uptriangle{OliveGreen}~\textbf{88.4} $\pm$ \textbf{0.4} & 80.9 $\pm$ 0.5\\
 & Radial ensemble & \uptriangle{OliveGreen}~79.4 $\pm$ 1.0 & \uptriangle{OliveGreen}~77.1 $\pm$ 0.9 & 66.2 $\pm$ 1.0 & \uptriangle{OliveGreen}~89.1 $\pm$ 0.5 & \uptriangle{OliveGreen}~87.4 $\pm$ 0.5 & 80.2 $\pm$ 0.6\\\hline
\multirow{5}{*}{Messidor-2} & Deep ensemble & \uptriangle{OliveGreen}~92.0 $\pm$ 0.8 & \uptriangle{OliveGreen}~90.0 $\pm$ 0.9 & 81.4 $\pm$ 1.2 & \downtriangle{BrickRed}~65.0 $\pm$ 1.8 & \uptriangle{OliveGreen}~67.1 $\pm$ 1.4 & 63.7 $\pm$ 1.5\\
 & MC dropout ensemble & \uptriangle{OliveGreen}~\textbf{94.3} $\pm$ \textbf{0.7} & \uptriangle{OliveGreen}~91.4 $\pm$ 0.8 & 80.3 $\pm$ 1.3 & \downtriangle{BrickRed}~\textbf{72.9} $\pm$ \textbf{1.9} & \uptriangle{OliveGreen}~\textbf{73.6} $\pm$ \textbf{1.4} & \textbf{68.0} $\pm$ \textbf{1.5}\\
 & MFVI ensemble & \uptriangle{OliveGreen}~\textbf{94.3} $\pm$ \textbf{0.8} & \uptriangle{OliveGreen}~\textbf{92.1} $\pm$ \textbf{0.8} & \textbf{83.9} $\pm$ \textbf{1.2} & \downtriangle{BrickRed}~57.3 $\pm$ 2.2 & \uptriangle{OliveGreen}~61.8 $\pm$ 1.7 & 59.9 $\pm$ 1.5\\
 & GVI ensemble & \uptriangle{OliveGreen}~92.4 $\pm$ 0.9 & \uptriangle{OliveGreen}~90.3 $\pm$ 0.8 & 80.0 $\pm$ 1.4 & \downtriangle{BrickRed}~50.9 $\pm$ 2.5 & \downtriangle{BrickRed}~58.1 $\pm$ 1.9 & 58.4 $\pm$ 1.4\\
 & Radial ensemble & \uptriangle{OliveGreen}~93.9 $\pm$ 0.8 & \uptriangle{OliveGreen}~91.7 $\pm$ 0.8 & 82.2 $\pm$ 1.3 & \downtriangle{BrickRed}~61.9 $\pm$ 2.3 & \uptriangle{OliveGreen}~65.4 $\pm$ 1.6 & 63.1 $\pm$ 1.5\\\hline
\multirow{5}{*}{APTOS} & Deep ensemble & \downtriangle{BrickRed}~62.3 $\pm$ 3.0 & \downtriangle{BrickRed}~83.2 $\pm$ 0.6 & 85.4 $\pm$ 0.5 & \downtriangle{BrickRed}~33.7 $\pm$ 6.4 & \downtriangle{BrickRed}~\textbf{69.6} $\pm$ \textbf{1.1} & \textbf{71.0} $\pm$ \textbf{0.7}\\
 & MC dropout ensemble & \downtriangle{BrickRed}~66.4 $\pm$ 3.3 & \downtriangle{BrickRed}~\textbf{85.6} $\pm$ \textbf{0.6} & \textbf{87.9} $\pm$ \textbf{0.4} & \downtriangle{BrickRed}~19.5 $\pm$ 1.1 & \downtriangle{BrickRed}~46.1 $\pm$ 1.2 & 53.0 $\pm$ 1.0\\
 & MFVI ensemble & \downtriangle{BrickRed}~62.6 $\pm$ 3.1 & \downtriangle{BrickRed}~82.1 $\pm$ 0.7 & 84.3 $\pm$ 0.6 & \downtriangle{BrickRed}~39.6 $\pm$ 4.6 & \uptriangle{OliveGreen}~61.2 $\pm$ 1.1 & 60.7 $\pm$ 0.9\\
 & GVI ensemble & \downtriangle{BrickRed}~\textbf{66.9} $\pm$ \textbf{2.8} & \downtriangle{BrickRed}~84.7 $\pm$ 0.6 & 87.3 $\pm$ 0.5 & \downtriangle{BrickRed}~\textbf{45.2} $\pm$ \textbf{4.1} & \downtriangle{BrickRed}~58.0 $\pm$ 1.0 & 59.2 $\pm$ 0.8\\
 & Radial ensemble & \downtriangle{BrickRed}~56.9 $\pm$ 4.4 & \downtriangle{BrickRed}~83.3 $\pm$ 0.7 & 87.1 $\pm$ 0.4 & \downtriangle{BrickRed}~29.3 $\pm$ 4.8 & \downtriangle{BrickRed}~54.1 $\pm$ 1.2 & 60.6 $\pm$ 0.9\\

\end{tabular} 
} 
}
\caption{PIRC results using QWK-Risk as the uncertainty measure. Mean and standard deviation computed for 100 bootstrap resamples of the test data. Relative improvement to previous referral level is denoted with a green up-pointing triangle, orange equals sign, or red down-pointing triangle for increasing, equal, or worse performance, respectively.}
\label{tab:pirc_qwkrisk}
\end{table}

\section{Conclusions}
We have replicated the usefulness of entropy as uncertainty estimation in RDR classification task for most robust neural networks when training with the EyePACS benchmark dataset, and demonstrated to a smaller extend the same finding for the KSSHP clinical hospital dataset. We also observe that the quality of the uncertainty estimates on out-of-distribution tests decreases when the models are trained with the KSSHP set. In addition, we show that entropy is less suitable as a measure of uncertainty in the clinical 5-class PIRC classification task for the EyePACS and KSSHP datasets. We have proposed and demonstrated that an uncertainty measure based on the classifier risk using the quadratic weighted Cohen's kappa provides a useful measure for the within-distribution uncertainty quantification that improves uncertainty based retained performance in comparison to entropy. However, the QWK-Risk decreases the out-of-distribution quality of the uncertainty estimates for the Messidor-2 and APTOS datasets.

From clinical perspective, the classifiers should be able to indicate uncertainty, such that the cases for which the classifier is confident can be automatically classified, while the difficult cases can be manually verified, and possibly corrected, by medical experts. Since most of the benchmark datasets only permit research use, for real world use-cases the classifiers should be able to be trained using "in-house" hospital datasets. When training the classifier on a hospital dataset, we expect that it can be utilized for future cases within the same hospital. In our studies we did not observe competitive performance for the KSSHP set on the clinical PIRC system, when entropy was utilized to quantify uncertainty. This highlights the concern that methods developed using benchmark datasets might not generalize to the clinical setting. However, the proposed QWK-Risk uncertainty measure enabled the models to surpass the state-of-the-art when 30\% of the most uncertain cases were referred, which demonstrates that the robust neural networks with a more appropriate uncertainty function may be utilized for clinical datasets and clinical use-cases.

The degradation and variability in out-of-distribution performance may partly be caused by different grading conventions and other grading variability may prevent correct evaluation. There are several mechanisms causing variable grading. Firstly, the frequency of photographic screening may affect the distribution of retinopathy severity with advanced grades of retinopathy being rarer when screening is more frequent. The patients are referred and treated before the advanced stages and followed by clinical visits instead of fundus photography. Secondly, there are no grading schemes available for the classification of treatment outcomes after the treatment. As a result, severe retinopathy may be arbitrarily classified into many different grades. The third reason for variable grading is poor quality of the retinal images. The grading of poor quality photographs may reflect some preconceived notions of patient status, based on such factors as age and type of diabetes, availability of treatments and also previous treatments.

Our future work includes a more fine-grained analysis of the out-of-distribution performance of robust neural networks. We will be evaluating a larger dataset of diabetic retinopathy images from the Helsinki region in Finland. In addition to the country distribution shift, our aim is to examine the within country hospital region distribution shift using this data, in addition to the KSSHP dataset. Additionally, our future work includes utilization of some of the computationally more intensive solutions for robust deep learning. Moreover, we will be examining the impact of joint training from the different regions on performance and robustness. Lastly, we aim to examine multi modal inputs with the fusion of multi-view retinal images with the task of even finer grain diabetic retinopathy grading.

\section{Methods}

\subsection{Datasets}
All our datasets consist of color images of the human retina and are graded using the following 5-class PIRC system for the severity scheme of diabetic retinopathy: \textit{no diabetic retinopathy} (class 0), \textit{mild diabetic retinopathy} (class 1), \textit{moderate diabetic retinopathy} (class 2), \textit{severe diabetic retinopathy} (class 3), and \textit{proliferative diabetic retinopathy} (class 4). From the PIRC system, we derive the binary \textit{referable/non-referable diabetic retinopathy} (RDR) classification, which is defined as the union of no diabetic retinopathy or mild diabetic retinopathy ($\leq1$) and referable as retinopathy worse than or equal to moderate diabetic retinopathy ($\geq2$). We chose to include the binary RDR system for comparison with the previous works.

The division of data to training, validation, and test sets were performed for EyePACS and KSSHP image datasets. For the EyePACS dataset, the official training set was used for training, the "Public" set was used for validation, and the "Private" set was used as the test set. For the KSSHP dataset, we used 70\%, 10\%, and 20\% of images for the training, validation, and test sets, respectively. For the splits, we used stratified pseudo random sampling to preserve PIRC class distribution of the original set, but taking into account that images from the same patient cannot reside in more than one set. Due to the low amount of images in the APTOS and Messidor-2 datasets, 3662 and 1744 respectively, they were used only as out-of-distribution test sets. Complete description of the training, validation, and test sets are described in Table \ref{datasets} and the test set class distributions in Table \ref{testdist}. The images were resized into standard size 512 x 512 and, to reduce known variability, preprocessed with steps described in the supplement.

\begin{table}[h]
    \centering
    \begin{tabular}{l|l|l|l|l}
        Subset&EyePACS\citep{eyepacs}&KSSHP\citep{retinaduodecim}&APTOS\citep{aptosdataset}&Messidor-2\citep{messidorcite2}\\\hline
        Train & 35125 (39.6\%) & 39482 (70.0\%)& - & -\\
        Validation & 10906 (12.3\%)& 5652 (10.0\%)& - & -\\
        Test & 42669 (48.1\%)& 11285 (20.0\%)& 3662 (100\%)& 1744 (100\%)\\ \hline
        Total & 88700 & 56419 & 3662 & 1744        
    \end{tabular}
    \caption{Number of images in each subset for each dataset.}
    \label{datasets}
\end{table}

\begin{table}[h]
    \centering
    \begin{tabular}{l|l|l|l|l}
Class & EyePACS\citep{eyepacs} & KSSHP \citep{retinaduodecim} & APTOS\citep{aptosdataset} & Messidor-2\citep{messidorcite2} \\ \hline
PIRC 0 & 31403 (73.6\%) & 7723 (68.4\%) & 1805 (49.3\%) & 1017 (58.3\%) \\
PIRC 1 & 3042 (7.1\%) & 2431 (21.5\%) & 370 (10.1\%) & 270 (15.5\%) \\
PIRC 2 & 6281 (14.7\%) & 930 (8.2\%) & 999 (27.3\%) & 347 (19.9\%) \\
PIRC 3 & 977 (2.3\%) & 177 (1.6\%) & 193 (5.3\%) & 75 (4.3\%) \\
PIRC 4 & 966 (2.3\%) & 24 (0.2\%) & 295 (8.1\%) & 35 (2\%) \\ \hline
RDR 0 & 34445 (80.7\%) & 10154 (90.0\%) & 2175 (59.4\%) & 1279 (73.3\%) \\
RDR 1 & 8224 (19.3\%) & 1131 (10.0\%) & 1487 (40.6\%) & 465 (26.7\%) \\
    \end{tabular}
    \caption{Class distribution for PIRC and RDR classification schemes of the test sets.}
    \label{testdist}
\end{table}

\subsection{Approximate Bayesian Deep Learning}
The approximate Bayesian deep learning models take the uncertainty in account by computing the following posterior predictive distribution:
\begin{equation}\label{ppeq}
    p(y\mid \boldsymbol{x}, D) = \int_{\boldsymbol{\theta} \in \Theta} p(y \mid \boldsymbol{x}, \boldsymbol{\theta}) ~ p(\boldsymbol{\theta} \mid D) ~ d\boldsymbol{\theta}.
\end{equation}
Here $\boldsymbol{x}$ and $y$ denote the image and target, respectively, $D$ the training data, and $\boldsymbol{\theta}$ the model parameters. The predictions are weighted averages using the posterior distribution $p(\boldsymbol{\theta} \mid D)$ \citep{vehtari,bbb}. 
For the standard neural networks, the maximum likelihood (ML) or maximum a posteriori (MAP) estimates are typically used, which completely ignore the uncertainty in the parameters.

The exact solution for Equation \eqref{ppeq} is intractable for deep neural networks, and Markov Chain Monte Carlo is prohibitively expensive for real world scenarios. Thus approximations need to be utilized. We selected the deep ensemble, MC dropout, MFVI, GVI, and Radial BNN as our approximate Bayesian methods. The posterior predictive distribution can be inexpensively approximated with these methods using an approximate posterior distribution $q(\boldsymbol{\theta})$ and Monte Carlo integation with $N$ samples:
\begin{equation}\label{mcppeq}
    p(y\mid \boldsymbol{x}, D) \approx \frac{1}{N}\sum_{i=1}^N p(y \mid \boldsymbol{x}, \boldsymbol{\theta}_i), ~~\boldsymbol{\theta}_i\sim q(\boldsymbol{\theta}).
\end{equation}

For all our experiments, we use a VGG16 type network as the base architecture identical to \citep{radial}. Our baseline network is a MAP solution, which was regularized with L2 weight decay and dropout. Model implementation and optimization are described in detail in the Supplementary Section \ref{supp_methods}.

\subsection{Deep Ensembles}
The deep ensemble is a collection of multiple ML or MAP neural network models. The models are trained using different initializations in order to produce a diverse set of models. The predictions of the models are averaged to produce the prediction of the ensemble model, corresponding to heuristically setting the posterior as uniform distribution over the set if models in Equation \eqref{mcppeq}.

\subsection{Monte Carlo Dropout - MC Dropout}
The Monte Carlo dropout method introduced in \citet{mcdropout} allows approximate Bayesian inference during test-time for networks that have been trained using the dropout regularization method. The dropout works by sampling a binary mask $\boldsymbol{r} = [r_1,\dots,r_d]^\top$ from a Bernoulli distribution $r_i\sim \text{Bern}(p)$ and masks the activations $h$ of a layer by computing the Hadamard product $r \odot h$ \citep{dropout}, which is equivalent to masking rows of the weight matrix and elements of the bias vector \citep{mcdropout}. The posterior predictive distribution is then computed using the dropout distribution in Equation \eqref{mcppeq}.

\subsection{Mean Field Variational Inference - MFVI}
The Mean Field variational approximations assume that the neural network parameters are independent and typically that the approximate posterior and prior are Gaussians \citep{graves,systematic,radial}. The evidence lower-bound (ELBO) can then be maximized, which provides a lower bound for the posterior probability, and for the diagonal multivariate Gaussian case, the equations become simple \citep{graves}:
\begin{align}\label{elbo}
    \mathcal{L}_{ELBO}(\mathcal{D},\boldsymbol{\theta}) 
    &= \mathbb{E}_{q(\boldsymbol{\theta})}[\log(p(\mathcal{D}\mid\boldsymbol{\theta}))]
    - D_{KL}[q(\boldsymbol{\theta})\mid\mid p(\boldsymbol{\theta})] \\
    &= \mathbb{E}_{q(\boldsymbol{\theta})}[\log(p(\mathcal{D}\mid\boldsymbol{\theta}))]
    - \sum_{j=1}^J \log \frac{s_j}{\sigma_j} + \frac{1}{2s_j^2}[(\mu_j-m_j)^2 + \sigma_j^2 - s_j^2]. \nonumber
\end{align}
Here the $KL[\cdot\mid\mid\cdot]$ denotes the Kullback-Leibler (KL) divergence, the prior is $\mathcal{N}(m_j,s_j^2)$, and the variational posterior is $\mathcal{N}(\mu_j,\sigma_j^2)$ for a parameter $\theta_j$. The remaining expected log-likelihood term is computed using Monte Carlo integration and the reparametrization trick \citep{vae}. Similar to the MC dropout, the posterior predictive distribution is computed using samples from the fitted approximate posterior distribution. We used multivariate standard normal distribution as the prior in all experiments.

\subsection{Radial Bayesian Neural Networks
- Radial BNN}
The multivariate normal distribution has the so-called "soap bubble" pathology in high dimensions, meaning that most of the probability mass is concentrated on thin shell far from the mean, resulting in samples with a high norm. In \citet{radial}, the high norm is proposed to be the problem in training deep neural networks using the MFVI approach with Gaussian posteriors. To address this issue, they propose a novel "Radial BNN" posterior distribution. The proposed distribution is constructed such that the samples have the same expected norm as univariate standard normal distribution, regardless of the dimension. The sampling process is defined as follows for a single weight vector:
\begin{align}
    \boldsymbol{w} &= \boldsymbol{\mu} + \boldsymbol{\sigma}\odot\boldsymbol{\hat{\epsilon}}\cdot r, \\
    \boldsymbol{\hat{\epsilon}} &= \frac{\boldsymbol{\epsilon}}{||\boldsymbol{\epsilon}||_2}, \\
    \boldsymbol{\epsilon} &\sim \mathcal{N}(\boldsymbol{0},I), \\
    r &\sim \mathcal{N}(0,1).
\end{align}
The resulting random variable $\boldsymbol{w}$ avoids the soap bubble pathology, however, it has no closed form probability density function or KL-divergence. The authors observe that a stochastic estimate of the KL-divergence can be computed similar to \citep{bbb}, allowing for optimizing the ELBO up to a constant. Multivariate standard normal distribution was selected as the prior, similar to MFVI.

\subsection{Generalized Variational Inference - GVI}
In \citet{gvi}, a novel optimization-centric view to posterior inference is proposed. The problem of finding posterior distributions is viewed in terms of the so-called "Rule-of-Three", which separates the loss, divergence, and the space of feasible solutions as different aspects of the optimization procedure. Different configurations result in different posterior inference methods, for example the standard Bayesian posterior inference and variational inference. This view allows for principled use of divergences, which are robust to the mis-specification of the prior. Since the diagonal standard normal is typically chosen for computational convenience (rather than to incorporate any prior knowledge about good values of the neural network weights) \citep{gvi}, this approach is promising. We select the robust divergence as R\'enyi's $\alpha$-Divergence ($D_{AR}^\alpha[~\cdot \mid \mid \cdot~]$), with $\alpha = 0.5$. As the loss function, we use the negative log-likelihood, and as the space of feasible solutions the mean field normal posteriors and priors, where the prior was selected as the diagonal standard normal. The total minimization objective is then:
\begin{align}
    \mathcal{L}_{GVI} &= -\mathbb{E}_{q(\boldsymbol{\theta})}[\log(p(\mathcal{D}\mid\boldsymbol{\theta}))]
    + D_{AR}^{\alpha}[q(\boldsymbol{\theta})\mid\mid p(\boldsymbol{\theta})], \\
    D_{AR}^{\alpha}[q(\boldsymbol{\theta})\mid\mid p(\boldsymbol{\theta})] &= \frac{1}{\alpha(1-\alpha)}\log\int q(\boldsymbol{\theta})^\alpha p(\boldsymbol{\theta})^{(1-\alpha)}d\boldsymbol{\theta}.
\end{align}

\subsection{Uncertainty estimation}

Many uncertainty estimates have been used in previous works. The entropy of the posterior predictive distribution has been a typical choice, and has been observed to work well for the binary RDR classification. However, for the 5-class PIRC classification, we observe that the entropy does not systematically improve the referral process QWK on the clinical dataset. We seek to identify alternative measures of uncertainty, which would give more informed rejection rules.

We measure the performance of our 5-class PIRC classifiers using the quadratic weighted Cohen's kappa \citep{qwk}:
\begin{align}\label{eq:qwk}
    \kappa_{QW}(C) &= 1 - \frac{\sum_{i=1}^M\sum_{j=1}^M(i-j)^2C_{i,j}}{\sum_{i=1}^M\sum_{j=1}^M(i-j)^2E_{i,j}}, \\
    E_{i,j}&=\frac{1}{N}\sum_{a=1}^MC_{i,a}\sum_{b=1}^MC_{b,j}.
\end{align}
The $C$ is the confusion matrix where element $C_{i,j}$ is the number of cases where the predicted label is $i$ and the true label is $j$, and $E$ is the expected agreement matrix. For a perfect classifier the confusion matrix is diagonal and thus the numerator term is zero, which results in a QWK value of 1. The QWK weights the misclassifications with squared distance of the numerical class labels, as well as with the expected agreement.

The referral of uncertain examples is essentially a case of \textit{reject option classification} \citep{bishop}. In reject option classification, the risk of a classification decision is the measure of uncertainty, and when the risk exceeds a certain threshold, the prediction is discarded \citep{rejoption_optimal}. When the classifier risk is the minimum risk over the decisions, and the error is defined using the 0/1 loss, a classic result is that the prediction is rejected if $\max_i p(y = i \mid x)<\tau$ for some threshold $\tau$ \citep{classic_reject,bishop}.

Instead of the minimum risk estimator, we choose to use an average risk, similar to the expected risk of classifier in \citet{lossfunctions}. The average risk view reveals an interesting connection between the classifier risk analysis and the entropy referral process, and thus helps in the analysis of constructing alternative uncertainty measures. The expected risk of a classifier that estimates the likelihood of discrete labels $y$ given $\boldsymbol{x}$ is defined as \citep{lossfunctions}
\begin{equation}
    \mathcal{I} = \int_{\boldsymbol{x}} \sum_y \mathcal{L}(p(y \mid \boldsymbol{x}),y)p(\boldsymbol{x},y)d\boldsymbol{x}.
\end{equation}
The $\mathcal{L}(\cdot,\cdot)$ is the loss function associated with a certain prediction and label combination. The risk associated with using the classifier for a certain input $\boldsymbol{x}$ can be derived by leaving the marginalization over $\boldsymbol{x}$ out, which is the expected (over $y$) conditional (on $\boldsymbol{x}$) risk:
\begin{equation}
    \mathcal{R}(\boldsymbol{x}) = \sum_y \mathcal{L}(p(y \mid \boldsymbol{x}),y)p(y \mid \boldsymbol{x}).
\end{equation}
For a given classifier, the risk is now completely defined by the choice of the loss function. Indeed, the expected conditional risk can be used for any performance measure we want to optimize by choosing a loss function that corresponds to the performance measure.

The negative log-likelihood of a target label $c$ given a categorical distribution $p(y \mid \boldsymbol{x})$ with $M$ classes is:
\begin{align}
    \mathcal{L}_{NLL}(p(y \mid \boldsymbol{x}),c) &= -\sum_{j=1}^M[c=j]\log(p(y=j \mid \boldsymbol{x})),\\
    &= -\log(p(y=c \mid \boldsymbol{x})). \nonumber
\end{align}
When we plug this to the expected conditional risk, we obtain the entropy of the posterior predictive distribution:
\begin{align}
    \mathcal{R}(\boldsymbol{x}) &= \sum_{i=1}^M \mathcal{L}_{NLL}(p(y \mid \boldsymbol{x}),i)p(y=i \mid \boldsymbol{x}),\\
    &= \sum_{i=1}^M -\log(p(y=i \mid \boldsymbol{x}))p(y = i \mid \boldsymbol{x}). \nonumber
\end{align}
Thus the negative log-likelihood induces a risk measure that is the entropy of the posterior predictive distribution.

In order to apply the methodology to the QWK, we need a loss function that directly reflects it. For a single prediction, the numerator and denominator terms in Equation \eqref{eq:qwk} will be the same and thus the $\kappa_{QWK}$ will always be zero. Thus we need the confusion matrix to be a non single entry matrix. For this purpose, we utilize an initial estimate of the confusion matrix $C$, computed on the validation set of the corresponding training set the model was trained on, and define the confusion matrix as a sum of the initial confusion matrix and a single entry matrix $S_{j,i}$ with 1 on index $j,i$. The entry $j,i$ denotes a combination of a prediction-target pair where the predicted label is $j$ and the target label is $i$. We propose the loss to then be the negative expected QWK:
\begin{equation}
    \mathcal{L}_{QWK}(p(y \mid \boldsymbol{x}),i) = -\sum_{j=1}^M p(y=j \mid \boldsymbol{x})\kappa_{QW}(C + S_{j,i}).
\end{equation}
The final QWK-Risk uncertainty estimate is then:
\begin{equation}
    \mathcal{R}_{QWK}(\boldsymbol{x}) = -\sum_{i=1}^M p(y=i \mid \boldsymbol{x})\sum_{j=1}^M p(y=j \mid \boldsymbol{x})\kappa_{QW}(C + S_{j,i}).
\end{equation}
The QWK-Risk can be interpreted as the expected negative QWK value for an input example $\boldsymbol{x}$, similar to entropy being the expected negative log-likelihood. The QWK-Risk can then be directly used in the referral process.

\section{Data availability}
The EyePACS dataset is available through the Kaggle competition in \url{https://www.kaggle.com/c/diabetic-retinopathy-detection} and the  APTOS dataset in \url{https://www.kaggle.com/c/aptos2019-blindness-detection}. The Messidor-2 dataset was kindly provided by the Messidor program partners (see https://www.adcis.net/en/third-party/messidor/), dataset available through \url{https://www.adcis.net/en/third-party/messidor2/}, and the Messidor-2 PIRC annotations are available in \url{https://www.kaggle.com/google-brain/messidor2-dr-grades}.

The KSSHP dataset is a non-open dataset to which local and EU jurisdictional restrictions (such as GDPR) apply. Inquiries may be sent to the corresponding author.

\section{Code availability}
All the deep learning implementations are available through publicly available repositories, The MAP networks and dropout networks were implemented using standard Pytorch \citep{pytorch} library functions. The diagonal multivariate normal distributions, used for MFVI and GVI, as well as the Kullback-Leibler divergence, were implemented using Pytorch distributions package \textit{torch.distributions}, available through the main Pytorch package. The R\'enyi's $\alpha$-Divergence was implemented according to the description in \citet{gvi}, and is presented in the Supplementary Section \ref{supp_methods}. The Radial BNN was reimplemented following the accompanying code for the seminal paper \citep{radial}, that also includes code for the VGG-style network architecture used in this study.

\bibliographystyle{IEEEtranN}
\bibliography{ref}

\end{document}